\documentclass{article}

\usepackage{microtype}
\usepackage{graphicx}
\usepackage{booktabs} %

\usepackage{amsfonts} %
\usepackage{nicefrac} %
\usepackage{microtype} %
\usepackage{gensymb}
\usepackage{subcaption}
\usepackage{amsthm}
\usepackage{mathrsfs}
\usepackage{amsmath}
\usepackage{amssymb}
\usepackage{mathtools}
\usepackage{wrapfig}
\usepackage{soul}
\usepackage{multirow}

\usepackage{hyperref}

\usepackage[accepted]{icml2020}

\usepackage{algorithmic}

\usepackage{hhline}

\newcommand{\E}{\mathbb{E}}

\newcommand{\phiv}{\boldsymbol{\phi}}
\newcommand{\ba}[1]{\begin{align}#1\end{align}}

\newcommand{\cdotv}{\boldsymbol{\cdot}}

\DeclareMathOperator*{\argmax}{argmax}

\usepackage{stackengine}
\newcommand\ubar[1]{\stackunder[1.2pt]{$#1$}{\rule{1.2ex}{.09ex}}}
\makeatletter
\newcommand{\distas}[1]{\mathbin{\overset{#1}{\kern\z@\sim}}}%

\newcommand{\cA}{\mathcal{A}}
\newcommand{\cL}{\mathcal{L}}

\newcommand{\cN}{\mathcal{N}}

\newcommand{\beqs}{\vspace{0mm}\begin{eqnarray}}
\newcommand{\eeqs}{\vspace{0mm}\end{eqnarray}}
\newcommand{\barr}{\begin{array}}
\newcommand{\earr}{\end{array}}

\newcommand{\Imat}{{\bf I}}

\newcommand{\hv}[0]{{\boldsymbol{h}} }

\newcommand{\mv}[0]{{\boldsymbol{m}}}

\newcommand{\rv}{\boldsymbol{r}}

\newcommand{\xv}{\boldsymbol{x}}

\newcommand{\zv}{\boldsymbol{z}}

\newcommand{\Sigmamat}[0]{{\boldsymbol{\Sigma}}}

\newcommand{\betav}[0]{{\boldsymbol{\beta}}}

\newcommand{\epsilonv}{\boldsymbol{\epsilon}}
\newcommand{\psiv}{\boldsymbol{\psi}}
\newcommand{\varepsilonv}{\boldsymbol{\varepsilon}}

\newcommand{\thetav}{\boldsymbol{\theta}}

\newcommand{\muv}[0]{{\boldsymbol{\mu}}}

\newcommand{\R}{\mathbb{R}}
\newcommand{\given}{\,|\,}

\icmltitlerunning{Thompson Sampling via Local Uncertainty}

\begin{document}

\twocolumn[
\icmltitle{Thompson Sampling via Local Uncertainty}

\begin{icmlauthorlist}
\icmlauthor{Zhendong Wang}{UT}
\icmlauthor{Mingyuan Zhou}{UT}
\end{icmlauthorlist}

\icmlaffiliation{UT}{McCombs School of Business, 
The University of Texas at Austin, Austin, TX 78712, USA}

\icmlcorrespondingauthor{Mingyuan Zhou}{mingyuan.zhou@mccombs.utexas.edu}

\icmlkeywords{Contextual bandits, variational inference, neural networks, latent variable models}

\vskip 0.3in
]

\printAffiliationsAndNotice{} %

\begin{abstract}
Thompson sampling is an efficient algorithm for sequential decision making, which exploits the posterior uncertainty to address the exploration-exploitation dilemma. There has been significant recent interest in integrating Bayesian neural networks into Thompson sampling. Most of these methods rely on global variable uncertainty for exploration. In this paper, we propose a new probabilistic modeling framework for Thompson sampling, where local latent variable uncertainty is used to sample the mean reward. Variational inference is used to approximate the posterior of the local variable, and semi-implicit structure is further introduced to enhance its expressiveness. Our experimental results on eight contextual bandit benchmark datasets show that Thompson sampling guided by local uncertainty achieves state-of-the-art performance while having low computational complexity.
\end{abstract}

\section{Introduction}
There has been significant recent interest in employing
deep neural networks %
to better solve sequential decision-making problems, such as these in reinforcement learning \citep{mnih2013playing,mnih2015human,mnih2016asynchronous,arulkumaran2017deep,franccois2018introduction} and contextual bandits \citep{riquelme2018deep, russo2018tutorial}. %
In a typical setting, 
by sequentially interacting with the environment, the agent or algorithm needs to learn how to take a sequence of decisions in order to maximize the expected cumulative reward. 
These problems are frequently encountered in various practical applications, ranging from clinical trials to recommender systems to anomaly detection \citep{djallel2019}.
Using a deep neural network as a powerful function approximator, whose task is to learn the mapping from an observed contextual feature vector to the hidden reward distributions, has become a common practice. 
Since the model training and data collection usually happen at the same time, the model needs to not only accurately approximate the distribution of the observed data, but also gain enough flexibility to predict that of the future data.

Addressing the exploration-exploitation dilemma %
is a vital part of sequential decision making. %
To maximize the expected cumulative reward, the agent needs to balance its effort in exploration, which chooses actions that may potentially increase its %
understanding of the environment, %
and its effort in exploitation, which takes the action that is expected to be the best given existing information. Typically, under-exploration will possibly trap the agent at a bad local optimal solution, %
while over-exploration could lead to a significant exploration cost. 
Various strategies have been proposed to tackle the exploration-exploitation dilemma, such as $\epsilon$-greedy \citep{ sutton1998introduction}, upper-confidence bound
\citep{auer2002uc}, Boltzmann exploration
\citep{cesa2017be, sutton1998introduction}, and Thompson sampling
\citep{thompson1933likelihood}. More recently, carefully adding random noise to model parameters
\citep{plappert2017ps, Fortunato2017nn, gal2016da} or bootstrap sampling \citep{osband2016de} before decision making also provide effective ways to encourage exploration. 

Thompson sampling (TS) \citep{thompson1933likelihood}, an elegant and widely used exploration strategy, is known for both its simplicity and good practical performance \citep{chapelle2011empirical,agrawal2012analysis,agrawal2013ts, russo2018tutorial}. TS will keep updating the posteriors of the parameters of the hidden reward distributions, and take actions according to the posterior predictive distributions of the rewards. 
Relying on posterior uncertainty to do exploration is the promising point of TS. Unfortunately, the exact posteriors are tractable for only a few models with limited representation power. Therefore, significant effort has been dedicated to posterior approximation. A recent development along this direction is empowering TS with Bayesian neural networks \citep{hinton1993keeping,bishop2006pattern,graves2011practical,neal2012bayesian, hernandez2015probabilistic}, and relying on the posteriors of the neural network weights to perform exploration under the TS framework \citep{riquelme2018deep}. 

Various inference methods have been employed to capture the uncertainty of the neural network weights, including Bayes by backprop (BBB) \citep{charles2015bbb}, stochastic gradient Markov chain Monte Carlo (MCMC) \citep{welling2011bayesian,li2016psg, mandt2016ava}, point estimate combined with random sampling \citep{riquelme2018deep}, and interactive particles based approximation \citep{ruiyi2019ts}. All these methods are focused on modeling the uncertainty of the global variables ($i.e.$, weights of the deep neural network) to maintain flexible approximation to the posterior predictive distributions. 

In this paper, differing from all aforementioned methods, we propose to sample from a local latent variable distribution to model the uncertainty of the mean rewards of actions given the contextual input. Our framework uses a latent variable model to model the reward distribution given a contextual input, and encodes this input to approximate the posterior distribution of the local latent variable given both the context, which has already been observed, and reward, which is yet to be observed. To further improve the expressiveness of the latent distribution, we introduce a semi-implicit variational distribution structure into the framework. We test our framework on contextual bandits, a classical task in sequential decision making, to verify its effectiveness. Experimental results show that the proposed local uncertainty guided TS algorithms achieve state-of-the-art performance, while having low computational complexity. 

\section{TS via Global Uncertainty}

Below we first briefly review contextual bandits and TS.

\subsection{Contextual Bandits and TS}

In a contextual bandit problem, we denote $\xv \in \R^d$ as the $d$ dimensional context (state) given by the environment, $a \in \cA = \{1, \dots, C\}$ as the action in a finite discrete space of size $C$, and $r \in \R$ as the scalar reward. Commonly, the agent will interact with the environment sequentially for $T$ times. At each time $t = 1, \dots, T$, the agent observes a new context $\xv_t$, chooses action $a_t\in\mathcal A$ based on the information provided by $\xv_t$, and receives reward $r_t$ provided by the environment. The reward $r_t$ can be a deterministic mapping $r_t = f(\xv_t, a_t)$ or a more complicated stochastic mapping 
$r_t = f(\xv_t, a_t,\epsilonv_t)$, where $\epsilonv_t$ represents random noise. 
The interactions $(\xv_t, a_t, r_t)$ at different times are independent from each other. The objective of the agent is to maximize the expected cumulative reward $\E [\sum_{t=1}^T r_t ]$, or equivalently to minimize the expected cumulative regret as 
\begin{equation}
 \text{CR}(T) = \sum_{t=1}^T \E \left[\max_{ a_t \in \cA} \E[ r(\xv_t, a_t)] - r_t\right].
\end{equation}

TS is a widely-used classical algorithm that has been shown to be effective for bandit problems both in practice \citep{chapelle2011empirical} and theory \citep{agrawal2012analysis}. Unlike the $\epsilon$-greedy algorithm that use parameter $\epsilon$ to control exploration and upper-confidence bound (UCB) that uses variance approximation to encourage exploration \citep{sutton1998introduction}, TS uses the uncertainty from the posterior samples of the model parameters for solving exploration-exploitation dilemma. If the model is not confident about its parameters, there will be large variations among the posterior samples, which will force the model to explore more to help better approximate the true underlying distribution. 

For contextual bandits, vanilla TS maintains a posterior distribution $p_t(\thetav)$ for global model parameter $\thetav$. For $t=1, \dots, T$, it samples $\thetav$ from its current posterior $p_{t-1}(\thetav)$ and uses the sampled $\thetav$ to transform the context-action pairs $(\xv_t, a)$ to estimate the mean rewards with $\hat{r}( a)= f(\xv_t, a;\thetav)$; after that, it greedily chooses the best action $a_t=\argmax_{ a\in \mathcal{A}}\hat{r}( a)$, receives reward $r_t$ from the environment, and uses the observed data to update its posterior on $\thetav$ via Bayes' rule. 
Vanilla TS faces the difficulty of balancing the complexity of the mapping function $f$ and tractability of posterior inference for $\thetav$, as discussed below.

\subsection{ Existing Global Uncertainty Guided TS}

In this section, we describe representative TS based algorithms for contextual bandits, which all share the same strategy of relying on the uncertainty of the global parameters ($e.g.$, neural network weights) that are shared across all observations to perform %
exploration. They differ from each other on how complex the mapping function $f$ is and how the posterior inference on $\thetav$ is implemented.

\textbf{Linear Method:} This method uses 
Bayesian linear regression with closed-form Gibbs sampling update equations, which relies on the posteriors of the regression coefficients for TS updates and maintains computational efficiency due to the use of conjugate priors. 
It assumes that at time $t$, the reward $y_t$ of an action given contextual input $\xv_t$ is generated as $y_t = \xv_t^T \betav + \epsilon_t$, 
where
$\betav$ is the vector of regression coefficients and $\epsilon_t \sim \cN (0, \sigma^2)$ is the noise. Note to avoid cluttered notation, here we omit the action index. This method places a normal prior on $\betav$ and inverse gamma prior on $\sigma^2$. At time $t$, given $\xv_t$ and the current random sample of $\betav$, it takes the best action under TS and receives reward $y_t$; with $\xv_{1:t}$ and $y_{1:t}$, it samples $\sigma^2$ from its inverse gamma distributed conditional posterior, and then samples $\betav$ from its Gaussian distributed conditional posterior; it proceeds to the next time and repeats the same update scheme under TS.

While this linear method accurately captures the posterior uncertainty of the global parameters $\betav$ and $\sigma^2$, its representation power is limited by both the linear mean and Gaussian distribution assumptions on reward $y$ given context $\xv$.
In practice,
the linear method often provides surprisingly competitive results, thanks to its ability to provide accurate uncertainty estimation.
 However, when its assumptions do not hold well in practice, such as when there are complex nonlinear dependencies between the rewards and contextual vectors, the linear method, even though with accurate posterior estimation, may not be able to converge to a good local optimal solution. Following \citet{riquelme2018deep}, we refer to this linear method as ``LinFullPost'' in what follows.

\textbf{Neural Linear:} To enhance the representation power of LinFullPost while maintaining closed-form posterior sampling, \citet{riquelme2018deep} propose the ``Neural Linear'' method, which feeds the representation of the last layer of a neural network as the covariates of a Bayesian linear regression model. It models the reward distribution of an action conditioning on $\xv$ as $y\sim\mathcal{N}(\betav^T\zv_{\xv},\sigma^2)$, 
where $\zv_{\xv}$ is the output of the neural network given $\xv$ as the input. It separates representation learning and uncertainty estimation into two parts. The neural network part is responsible for finding a good representation of $\xv$, 
while the Bayesian linear regression part is responsible for obtaining uncertainty estimation on $\betav$ and making the decision on which action to choose under TS. The training for the two parts can be performed at different time-scales. It is reasonable to update the Bayesian linear regression part as soon as a new data arrives, while to update the neural network part only after collecting a sufficient number of new data points. 

As Neural Linear transforms context $\xv$ into latent space $\zv$ via a deterministic neural network, the model uncertainty still all comes from sampling the global parameters $\betav$ and $\sigma$ from their posteriors under the Bayesian linear regression part. Hence, this method relies on the uncertainty of global model parameters to perform TS. 

\textbf{Bayes By Backprop (BBB):} This method uses variational inference to perform uncertainty estimation on the neural network weights \citep{charles2015bbb}. In order to exploit the reparameterization trick for tractable variational inference \citep{kingma2013vae}, it models the neural network weights with independent Gaussian distributions, whose means and variances become the network parameters to be optimized. However, 
the fully factorized mean-field variational inference used by BBB is well-known to have the tendency to underestimate posterior uncertainty \citep{jordan1999introduction,blei2017variational}.
Moreover,
it is also questionable whether the weight uncertainty can be effectively translated into reward uncertainty given context $\xv$ \citep{bishop2006pattern,shengyang2019fv}, especially considering that BBB makes both the independent and Gaussian assumptions on its network weights. For TS, underestimating uncertainty often leads to under exploration.
As the neural network weights are shared across all observations, 
BBB also relies on the uncertainty of global parameters to perform TS.

\textbf{Particle-Interactive TS via Discrete Gradient Flow %
($\pi$-TS-DGF):} The $\pi$-TS-DGF %
method of \citet{ruiyi2019ts} casts posterior approximation as a distribution optimization problem under a Wasserstein-gradient-flow framework. In this setting, posterior sampling in TS can be considered as a convex optimization problem on the space of probability measures.
For tractability, it maintains a set of particles that interact with each other and evolve over time to approximate the posterior. 
For contextual bandits, each particle corresponds to a set of neural network weights, and the algorithm uniformly at random chooses one particle at each time and uses it as a posterior sample of the neural network weights. 
A benefit of $\pi$-TS-DGF is that it imposes no explicit parametric assumption on the posterior distribution. 
However, it faces an uneasy choice of setting the number of particles. Maintaining a large number of particles means training many sets of neural network weights at the same time, which is considerably expensive in computation, while a small number might lead to bad uncertainty estimation due to inaccurate posterior approximation. The computational cost prevents $\pi$-TS-DGF from using large-size neural networks.
Similar to BBB, $\pi$-TS-DGF also relies on the uncertainty of global parameters to perform exploration.

\section{TS via Local Uncertainty}

Vanilla TS has several limitations. Its performance is sensitive to the accuracy of the mapping function $f(\xv, a; \thetav)$ and maintaining the exact posteriors for all model parameters is often %
infeasible.
Utilizing global parameter uncertainty to capture the %
posterior uncertainty of the mean rewards is challenging: first, the number of global parameters is often large, making it difficult to model their uncertainty under limited data without imposing strong assumptions; second, the model size is often constrained by the computational cost and training stability; third, the uncertainty on the global parameters %
may not be well translated into the uncertainty of the mean rewards by the mapping function.

To overcome these aforementioned limitations of vanilla TS, 
 we propose TS via local uncertainty (LU). Rather than following the convention to impose uncertainty on global parameter $\thetav$ to model the mean reward uncertainty, 
we apply uncertainty on local latent variables 
to balance exploration and exploitation
under TS. We first construct a neural network powered latent variable model to model the mean reward distribution, and then introduce a contextual variational distribution to model the pre-posterior uncertainty on the mean rewards, which is used to guide the selection of actions.
We first consider a contextual variational distribution using a diagonal Gaussian construction, and then another one using a semi-implicit construction.

\subsection{Local Variable based Mean Reward Estimation}

In a contextual bandit problem, the agent needs to continuously update its estimate %
of the unknown mean reward distributions through its interactions %
with the environment.
In this case, given context $\xv$, taking different actions $a$ will heavily impact the accumulated data for rewards, which predominantly influences the approximation of the mean reward distribution. 
TS via global uncertainty relies on the posterior sample of the global parameter to capture the uncertainty of the mean reward of an action at time $t$ as
\ba{\E[r_t\given \xv_t, a_t, \betav], ~~\betav\sim p(\betav\given \xv_{1:t-1}, a_{1:t-1},r_{1:t-1}). \label{eq:TS_meanreward}
}
It chooses the action whose mean reward given $\xv_t$ and $\betav$ is the largest, receives reward from the environment, and then updates the posterior of the global parameter $\betav$ before taking another contextual vector. 

By contrast, denoting $\rv_t\in \mathbb{R}^{|\mathcal A|}$ as the rewards of all actions in $\mathcal A$, we use a latent variable model to approximate the distribution of %
$\rv_t$ %
given $\xv_t$ as
$$ \rv_t \sim p(\rv_t \given \xv_t, \zv_t),~~\zv_t \sim p(\zv).$$
This provides a flexible marginal distribution, whose density is often intractable, to model $\rv_t$ given $\xv_t$ as
\begin{align}
p(\rv_t \given \xv_t) %
&= \E_{\zv_t \sim p(\zv )} [p(\rv_t \given \xv_t, \zv_t)].\label{LUmarginal}
\end{align}
To maximize the likelihood of this intractable marginal, we resort to variational inference \citep{jordan1999introduction,bishop2000vr, blei2017variational}. More specifically, related to auto-encoding variational Bayes \citep{kingma2013vae}, we introduce contextual variational distribution 
$q(\zv_t \given \xv_t)$ to approximate the posterior $p(\zv_t\given \xv_t,\rv_t)$ by minimizing the Kullback--Leiber (KL) divergence as
$\mbox{KL}(q(\zv_t \given \xv_t)||p(\zv_t\given \xv_t,\rv_t))$. 
Since one may show that $\log p(\rv_t\given \xv_t) = \cL_t + \text{KL} (q(\zv_t \given \xv_t) || p(\zv_t \given \xv_t,\rv_t))$ and the KL divergence is non-negative, where 
the evidence lower bound (ELBO) $\cL_t$ is expressed as
\begin{equation}
 \small 
 \textstyle \cL_t = \E_{\zv_t \sim q(\cdotv \given \xv_t)} \left[ \log p(\rv_t \given \xv_t, \zv_t) + \log \frac{p(\zv_t)}{q(\zv_t \given \xv_t)} \right], \label{eq:ELBOlocal}
\end{equation}
minimizing $\mbox{KL}(q(\zv_t \given \xv_t)||p(\zv_t\given \xv_t, \rv_t))$ becomes the same as maximizing the ELBO $\cL_t$.  We note if $p(\rv_t \given \xv_t, \zv_t)$ in \eqref{eq:ELBOlocal} is simplified as $p(\rv_t \given\zv_t)$, then  it becomes related to the variational hetero-encoder of \citet{Zhang2020Variational}.

Note for TS via LU,
distinct from the usual auto-encoding variational inference \citep{kingma2013vae}, we are facing an online learning problem, where we need to draw $\zv_t$ from variational posterior $q(\zv_t \given \xv_t)$ to estimate the actions’ mean rewards before we are able to choose an action and hence observe its reward. Moreover, we only observe the chosen action’s reward but not the other actions’. Therefore, the contextual variational distribution $q(\zv_t \given \xv_t)$ is not amortized over the actions' rewards at time $t$.
At time $t$, before optimizing the ELBO, we need to first sample from the mean reward distribution using
\ba{
\E[r_t\given \xv_t,a_t,\zv_t],~~%
~ \zv_t \sim q(\cdotv \given \xv_t),\label{eq:TS_Local_meanreward}
}
where $\E[r_t\given \xv,a_t,\zv_t] = \int r_t \cdot p(r_t\given \xv,a_t,\zv_t) dr_t$; we
then choose the action whose mean reward given $\zv_t$ and $\xv_t$ is the largest, %
observe the true reward $r_t$ returned by the environment, and optimize the parameters of $p$ and $q$ to maximize the ELBO in \eqref{eq:ELBOlocal}.  

Note a key difference between TS via LU and TS via global uncertainty is that to estimate the mean rewards of all actions, a random sample from the contextual variational distribution of the local variable $\zv_t$, as shown in \eqref{eq:TS_Local_meanreward}, has replaced the role of a random sample from the posterior distribution of the global variable $\betav$, as shown in 
\eqref{eq:TS_meanreward}. 
In other words, 
rather than approximating the posterior of global parameters, our model 
estimates the posterior of local latent variable $\zv_t$, and utilizes its uncertainty to perform exploration under TS. 
As global parameter $\betav$ often has a very high dimension ($e.g.$, when a neural network is used in $p(\rv_t \given \xv_t, \betav)$), one often has to impose strong assumptions ($e.g.$, independent Gauss with bounded variance) on its variational posterior for stable inference. By contrast, $\zv_t$ often has low dimension ($e.g.$, 50), which can be well modeled with flexible variational distribution. 
We use a neural network to define the deterministic mapping from 
$\xv_t$ and $\zv_t$ to the mean rewards of all actions, 
and train the network parameter according to the ELBO in \eqref{eq:ELBOlocal}. We describe two different versions of TS via LU, as will be discussed in detail, in %
Algorithms 2 and 3 in the Appendix,
 respectively.

In summary, the change from relying on global uncertainty to replying on local uncertainty brings several potential benefits. First, the neural network mapping $\xv_t$ and $\zv_t$ to the mean rewards can be made as complex as needed, without the need to worry about the tractability of posterior inference on the global parameters, the number of which is often so large that uncertainty estimation on them becomes possible only under strong distributional assumptions. Second, the uncertainty comes from the input feature space
rather than from the weight space, leading to more direct influence on the uncertainty of the mean rewards. Third, $\zv_t$ often has a much lower dimension than $\betav$, making it much more computationally efficient when the dimension of $\betav$ is high.

\subsection{Local Uncertainty Modeling with Gaussian Variational Posterior}\label{sec:GaussianLU}

We model both the rewards conditioning on $\xv_t$ and $\zv_t$ and the prior using diagonal Gaussian distributions as
\ba{p(\rv_t \given \xv_t, \zv_t) &= \cN (\muv_{\rv_t}, \Sigmamat_{\rv}),~\muv_{\rv_t} = \mathcal{T}_{\thetav} ([\xv_t,\zv_t]),\notag\\
&~~~~p(\zv_t) = \cN (0, \Sigmamat_{\zv}),\label{eq:decoder}}
where $\mathcal{T}_{\thetav}$ is a neural networks parameterized by $\thetav$ that maps the $[\xv_t, \zv_t]$ concatenation to the mean rewards of all actions as $\muv_{\rv_t}\in\mathbb{R}^{|\mathcal A|}$; both $\Sigmamat_{\rv} \in\mathbb{R}^{|\mathcal A|\times |\mathcal A|}$ and $\Sigmamat_{\zv}\in\mathbb{R}^{|\zv|\times |\zv|}$ are diagonal covariance matrix. Under this construction, the estimated mean rewards of all actions can be expressed as
\ba{
\E[\rv_t\given \xv_t,\zv_t] = \mathcal{T}_{\thetav}([\xv_t,\zv_t]). \label{eq:mapping}
}

We define the contextual variational distribution as a diagonal Gaussian distribution %
as
\ba{q(\zv_t \given \xv_t) &= \cN ( \muv_{\zv_t}, \Sigmamat_{\zv_t}),\notag\\
\muv_{\zv_t} = \mathcal{T}_{\phiv_1}(\hv),~~ \Sigmamat_{\zv_t} &= \mathcal{T}_{\phiv_2}(\hv),~~\hv = \mathcal{T}_{\phiv_0}(\xv_t),\label{eq:LU-Gauss}
}
where $\phiv_0$, $\phiv_1$, and $\phiv_2$ are neural network parameters. We describe how to address 
contextual bandit problems in Algorithm 2 in the Appendix, 
a limitation of which is that $q(\zv_t \given \xv_t)$ is restricted to be a Gaussian distribution with a diagonal covariance matrix, which might not be flexible enough to well model the true posterior 
that may exhibit multi-modality, skewness, heavy tails, and dependencies between different dimensions. To improve its expressiveness, below we leverage semi-implicit variational inference (SIVI) of \citet{mingzhang2018sivi} that mixes a distribution, which is simple to sample from but not required to be explicit, with an explicit and reparameterizable distribution to make the resulted hierarchical distribution more flexible, while maintaining tractable inference.

\subsection{Local Uncertainty Modeling with Semi-Implicit Variational Posterior}

Keeping likelihood $p(\rv_t \given \xv_t, \zv_t)$ and prior $p(\zv_t)$ the same as in \eqref{eq:decoder}, %
we model the contextual variational distribution using a semi-implicit construction as
\ba{
\textstyle q(\zv_t \given \xv_t) = \int q(\zv_t \given \psiv_t,\xv_t) q(\psiv_t\given \xv_t) d \psiv_t
\label{eq:SIVI}
}
where the first-layer explicit distribution is defined as
\ba{
q(\zv_t \given \psiv_t,\xv_t) = \mathcal N(\psiv_t, \Sigmamat_{\zv_t}),~\Sigmamat_{\zv_t} = \mathcal{T}_{\phiv_2}(\xv_t),\label{eq:LUSIVI1}
}
and the mean $\psiv_t$ is drawn from an implicit distribution, which generates its random samples by nonlinearly transforming random noise $\epsilonv_t\sim p(\epsilonv)$ as
\ba{
\psiv_t = \mathcal{T}_{\phiv_1}([\xv_t, \epsilonv_t]),~\epsilonv_t\sim p(\epsilonv). \label{eq:LUSIVI2}
}
We choose $p(\epsilonv)=\mathcal{N}(0,4\Imat)$ in this paper. Note while the probability density function of $p(\epsilonv)$ is analytic, that of $\psiv_t$ is implicit if the transformation $\mathcal{T}_{\phiv_1}$ is not invertible. 

While given $\epsilonv_t$ and $\xv_t$, the local latent variable $\zv_t$ follows a diagonal Gaussian distribution, the marginal distribution $q(\zv_t \given \xv_t)$, obtained by integrating out the random noise $\epsilonv_t$, becomes an implicit distribution that is no longer restricted to follow a diagonal Gaussian as in Section~\ref{sec:GaussianLU}. 
Thus, given the same mean reward mapping function as in \eqref{eq:mapping}, we can better capture the uncertainty on %
the mean rewards by sampling the local latent variable $\zv_t$ from a more flexible contextual variational distribution $q(\zv_t\given \xv_t)$ as in \eqref{eq:SIVI}.

While the original ELBO becomes intractable given an implicit contextual variational distribution, as in \citet{mingzhang2018sivi} and \citet{molchanov2019doubly}, we can optimize a lower bound of the ELBO that is amenable to direct optimization via stochastic gradient descent (SGD): 
we sample $K + 1$ $\psiv_t$'s, use only one of them to sample $\zv_t$ from the conditional distribution $q(\zv_t \given \psiv_t,\xv_t)$, %
and combine them for computing a lower bound of the ELBO as
$$
\begin{aligned}
 &\ubar{\cL}_{K,t} = \E_{\epsilonv_t^{(0)}, \dots, \epsilonv_t^{(K)}\stackrel{iid} \sim p({\epsilonv})} \E_{\zv_t \sim q(\cdotv \given \psiv_t^{(0)},\xv_t)} \\
 & \bigg[ \ln p(\rv_t \given \xv_t, \zv_t) + \ln \frac{p(\zv_t)}{\frac{1}{K+1}\sum_{k=0}^K q(\zv_t \given \psiv_t^{(k)},\xv_t)} \bigg],
\end{aligned}
$$
where $\psiv_t^{(k)}: = \mathcal{T}_{\phiv_1}([\xv_t,\epsilonv_t^{(k)}])$. % for $k=0,\ldots,K$.
Different from related works \citep{ranganath2016hierarchical,maaloe2016auxiliary} that also employ a hierarchical variational distribution, SIVI allows $q_{\phiv}(\psiv)$ to follow an implicit distribution \citep{huszar2017variational,tran2017hierarchical} and directly optimizes a surrogate ELBO.

We describe TS guided by semi-implicit LU in %
Algorithm~3 in the Appendix. The value of $K$ is related to how close $\ubar{\cL}_{K,t}$ is to $\cL_{t}$. A moderate value of $K=50$ is found to be sufficient for neural network training, which does not bring much extra computational cost. Benefiting from the expressiveness improvement of using the semi-implicit structure, the distribution of $\E[\rv_t \given \xv_t, \zv_t]$ under $q(\zv_t \given \xv_t)$ becomes more flexible and can fit more complicated mean reward distributions. While this added flexibility may %
slightly degrade the performance for problems with simple reward distributions, %
overall, semi-implicit local uncertainty is found to work better than Gaussian local uncertainty. %

\section{Experiments}

We evaluate Gaussian variational LU guided TS, referred to as LU-Gauss, and semi-implicit variational LU guided TS, referred to as LU-SIVI, on the contextual bandits benchmark used in 
\citet{riquelme2018deep}. 
We consider %
eight different datasets from this benchmark, including Mushroom, Financial, Statlog, Jester, Wheel, Covertype, Adult, and Census, which exhibit a wide variety of statistical properties. Details on these datasets are provided in 
Table \ref{tb:runtime}.
 For both LU-Gauss and LU-SIVI, we choose the Adam optimizer with the learning rate set as $10^{-3}$.
Python (TensorFlow 1.14) code for both LU-Gauss and LU-SIVI is available
at \url{https://github.com/Zhendong-Wang/Thompson-Sampling-via-Local-Uncertainty}

We compare the proposed LU-Gauss and LU-SIVI to LinFullPost, BBB, and Neural Linear implemented in \citet{riquelme2018deep}, and $\pi$-TS-DGF of \citet{ruiyi2019ts}.
The details on the neural network structures 
are provided in the Appendix. %
Since each experiment involves randomly sampling a subset of contextual vectors from %
the full dataset, all results %
are averaged over 50 independent random trials. In each random trial, we rerun the code of LinFullPost, BBB, and Neural Linear provided by \citet{riquelme2018deep} and the code of $\pi$-TS-DGF provided by \citet{ruiyi2019ts}.
To ensure a fair comparison, these 50 random sequences for each dataset are made the same across all the aforementioned algorithms. %
Note we have also considered making comparison with functional variational Bayesian neural networks (FBNNs) of \citet{shengyang2019fv}. However, as the code for FBNNs 
is too computationally expensive for us to run as many as 50 independent random trials for each of the eight benchmark datasets, we defer the details of an informal comparison with FBNNs to the Appendix.

\subsection{
Exploratory Analysis}\label{sec:exploratory}

To analyze how the performance is impacted by the expressiveness of the variational distribution $q(\zv_t \given \xv_t)$ that is used to model the local uncertainty, we choose the Mushroom dataset as an example, in which the stochastic rewards exhibit multi-modality. 
The Mushroom dataset has two distinct classes: Poisonous and Safe, and the agent has two possible actions: Eat or Not Eat. Eating a safe mushroom will be awarded $+5$, while eating a poisonous one will be awarded either $+5$ or $-35$, which are equally likely to occur. If the agent chooses to not eat the mushroom, it will receive $0$ reward. Thus, given a poisonous mushroom, the true reward distribution of Action Not Eat has a single mode at zero, while that of Action Eat has two modes, $+5$ and $-35$, with mean $-15$. For this reason, given a poisonous mushroom, a variational distribution that is not flexible enough will face the risk of concentrating the high density region of its mean reward distribution of Action Eat around $+5$, which is a local mode, leading to the wrong action; by contrast, a sufficiently flexible variational distribution could escape the local mode at $+5$ even if it initially concentrates its mean reward distribution around it, leading to better exploration.

Running both LU-Gauss and LU-SIVI on the same random sequence of contextual vectors selected from Mushroom, we perform two different exploratory analyses: 1) We %
pick one poisonous mushroom, whose contextual feature vector $\xv_p$ is being regarded as edible at the beginning of training by both LU-Gauss and LU-SIVI, and visualize how the
histogram of the sampled %
mean rewards %
of $\xv_p$ at each time step changes as the training progresses; 
2) We visualize how the %
histogram of the sampled 
mean rewards of all poisonous mushroom at each step changes as the training progresses.

\begin{figure*}[!t]
\centering
\begin{subfigure}{.22\textwidth}
 \centering
 \includegraphics[width=1\linewidth]{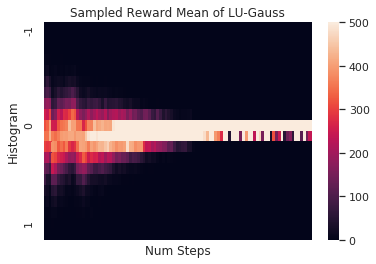}
 \vspace{-5mm}
 \label{fig:f11}
 \caption{LU-Gauss Not Eat}
\end{subfigure}
\begin{subfigure}{.22\textwidth}
 \centering
 \includegraphics[width=1\linewidth]{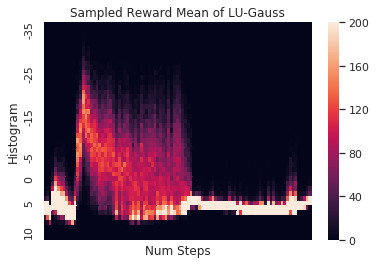}
 \vspace{-5mm}
 \label{fig:f12}
 \caption{LU-Gauss Eat}
\end{subfigure}
\begin{subfigure}{.22\textwidth}
 \centering
 \includegraphics[width=1\linewidth]{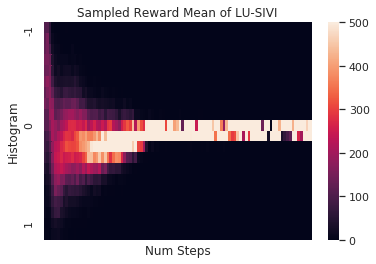} 
 \vspace{-5mm}
 \label{fig:f13}
 \caption{LU-SIVI Not Eat}
\end{subfigure}
\centering
\begin{subfigure}{.22\textwidth}
 \centering
 \includegraphics[width=1\linewidth]{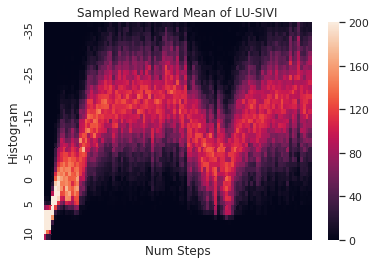} 
 \vspace{-5mm}
 \label{fig:14}
 \caption{LU-SIVI Eat}
\end{subfigure}
\vspace{-2.5mm}
\caption{The distribution of mean reward on a single poisonous mushroom \label{fig:singlemushroom}}
\vspace{-1mm}
\end{figure*}

More specifically, 
for the first analysis, for $s=1,\ldots,S$,
 we use \eqref{eq:LU-Gauss} to sample 
$\zv^{(s)}_{t,p}\given \xv_p \sim
q(\zv_{t,p}\given \xv_p)$ for LU-Gaussian, or use \eqref{eq:LUSIVI1} and \eqref{eq:LUSIVI2} to sample
$\zv^{(s)}_t\given \xv_p\sim
q(\zv_t\given \psiv_t^{(s)},\xv_p),~\psiv_t^{(s)}\given \xv_p \sim q(\psiv_t\given \xv_p)$ for LU-SIVI, and then use \eqref{eq:mapping} to compute a mean reward given $\xv_p$ and $\zv^{(s)}_{t,p}$ as $\E[\rv_{t,p}^{(s)}\given \xv_p,\zv_{t,p}^{(s)}] = \mathcal{T}_{\thetav}([\xv_p,\zv_{t,p}^{(s)}])$. We draw $S=2000$ mean reward samples 
for each $t$ to form a histogram for that training step, and visualize these histograms over training steps as a heatmap, as shown in Figure~\ref{fig:singlemushroom}.

\begin{figure*}[!t]
\centering
\begin{subfigure}{.22\textwidth}
 \centering
 \includegraphics[width=1\linewidth]{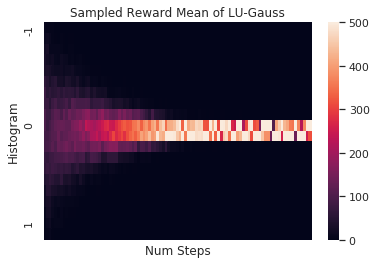}
 \vspace{-5mm}
 \label{fig:f21}
 \caption{LU-Gauss Not Eat}
\end{subfigure}
\begin{subfigure}{.22\textwidth}
 \centering
 \includegraphics[width=1\linewidth]{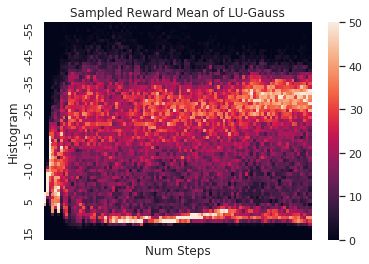}
 \vspace{-5mm}
 \label{fig:f22}
 \caption{LU-Gauss Eat}
\end{subfigure} 
\begin{subfigure}{.22\textwidth}
 \centering
 \includegraphics[width=1\linewidth]{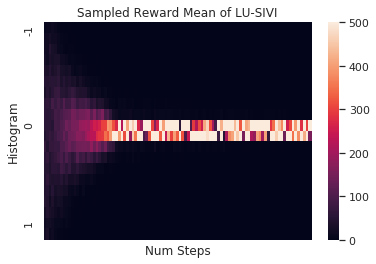} 
 \vspace{-5mm}
 \label{fig:f23}
 \caption{LU-SIVI Not Eat}
\end{subfigure}
\begin{subfigure}{.22\textwidth}
 \centering
 \includegraphics[width=1\linewidth]{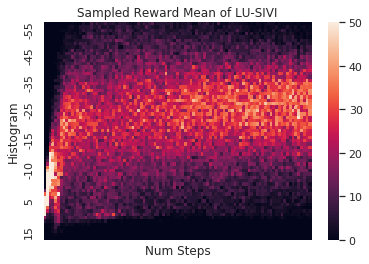}
 \vspace{-5mm}
 \label{fig:f24}
 \caption{LU-SIVI Eat}
\end{subfigure}
\vspace{-2.5mm}
\caption{Convergence of the distribution of mean reward on all poisonous mushrooms \label{fig:manymushroom}}
\vspace{-1mm}
\end{figure*}

For Action Not Eat, as shown in Figures~\ref{fig:singlemushroom} (a) and (c), the empirical distribution of the sampled mean rewards for this poisonous mushroom becomes more and more concentrated around zero %
for both LU-Gauss and LU-SIVI. %
For Action Eat, as shown in Figures~\ref{fig:singlemushroom} (b) and (d), while both
LU-Gauss and LU-SIVI initially consider this poisonous mushroom $\xv_p$ as edible with a positive reward, their mean reward distributions given $\xv_p$ become more and more different as the training progresses. In particular, LU-Gauss first becomes less certain and then more certain, with its high density region of the mean reward empirical distribution first shifting below zero, then moving back above zero, and eventually concentrating around $+5$, which means it mistakenly treats this poisonous mushroom as edible at many training steps. By contrast, LU-SIVI gradually increases its uncertainty and then maintains it around the same level, with its high density region of the mean reward empirical distribution quickly shifting below zero and remaining below zero at most of the training steps, which means it correctly identifies this poisonous mushroom most of the time.

For the second analysis, we take a single random sample of the mean reward for each mushroom at $t$, use the sampled mean rewards over all poisonous mushrooms as $\{\rv_{t,p}\}_{p}$ to form a histogram, and visualize the histograms over time as a heatmap, as shown in Figure~\ref{fig:manymushroom}.
For Action Not Eat, as shown in Figures~\ref{fig:manymushroom} (a) and (c), 
both LU-Gauss and LU-SIVI quickly capture the underlying reward distribution, concentrating the histograms around zero. 
For Action Eat, 
as shown in Figure~\ref{fig:manymushroom} (b), under LU-Gauss,
the sampled mean rewards of all poisonous mushrooms gradually concentrate around two density modes, with one clearly below zero and the other clearly above zero, suggesting that LU-Gauss will make a large number of mistakes in treating poisonous mushrooms as edible. By contrast, as shown in Figure~\ref{fig:manymushroom} (d), under LU-SIVI,
the sampled mean rewards of all poisonous mushrooms quickly move down their high density region below zero and then maintain a single density mode around $-30$, suggesting that LU-SIVI will correctly identify poisonous mushrooms most of the time. 

These analyses suggest that for the Mushroom dataset, whose true reward distribution of eating a poisonous mushroom exhibits two density modes, the variational distribution of LU-Guass shown in \eqref{eq:LU-Gauss} underperforms that of LU-SIVI shown in \eqref{eq:SIVI} in exploration and faces a greater risk to concentrate its mean reward distribution around the undesired local mode, leading to poorer performance. LU-SIVI, which introduces a more flexible varaitional distribution that is also amenable to optimization via SGD, achieves a better balance between exploration and exploitation.

\begin{figure*}[!t]
\centering
\begin{subfigure}{.24\textwidth}
 \centering
 \includegraphics[width=1\linewidth]{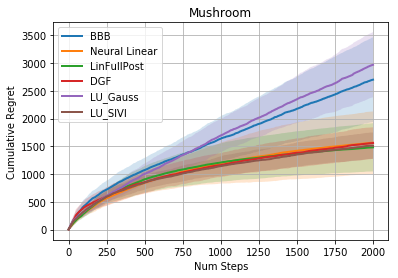}
 \vspace{-5mm}
 \label{fig:mushroom}
\end{subfigure}
\begin{subfigure}{.24\textwidth}
 \centering
 \includegraphics[width=1\linewidth]{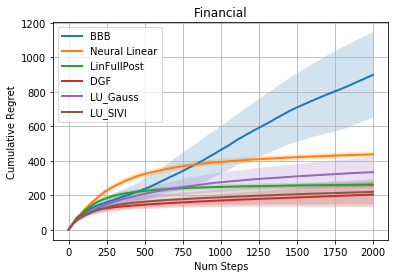}
 \vspace{-5mm}
 \label{fig:financial}
\end{subfigure} 
\begin{subfigure}{.24\textwidth}
 \centering
 \includegraphics[width=1\linewidth]{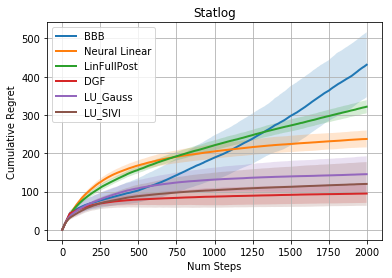} 
 \vspace{-5mm}
 \label{fig:statlog}
\end{subfigure}
\centering
\begin{subfigure}{.24\textwidth}
 \centering
 \includegraphics[width=1\linewidth]{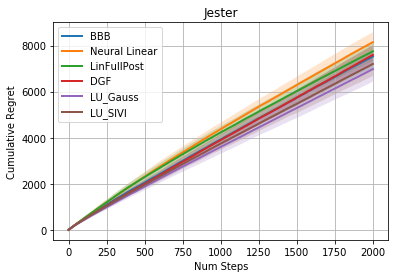} 
 \vspace{-5mm}
 \label{fig:jester}
\end{subfigure} \\
\begin{subfigure}{.24\textwidth}
 \centering
 \includegraphics[width=1\linewidth]{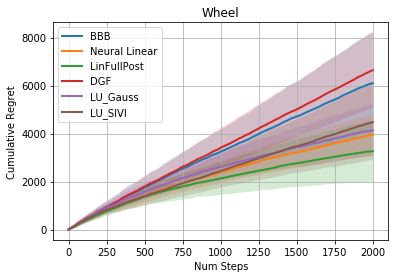} 
 \vspace{-5mm}
 \label{fig:wheel}
\end{subfigure}
\begin{subfigure}{.24\textwidth}
 \centering
 \includegraphics[width=1\linewidth]{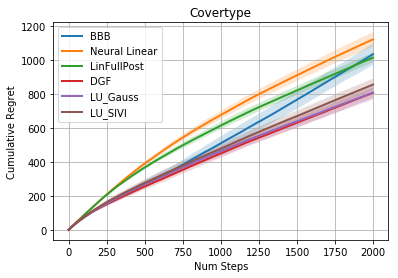}
\vspace{-5mm} 
\label{fig:covertype}
\end{subfigure} 
\begin{subfigure}{.24\textwidth}
 \centering
 \includegraphics[width=1\linewidth]{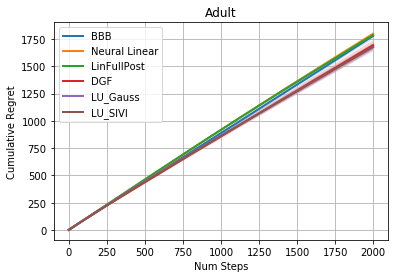}
\vspace{-5mm} 
\label{fig:adult}
\end{subfigure}
\begin{subfigure}{.24\textwidth}
 \centering
 \includegraphics[width=1\linewidth]{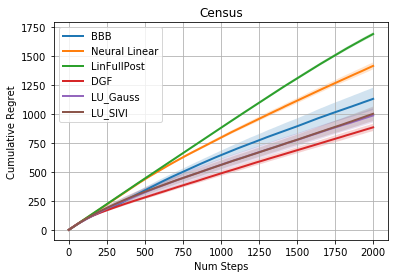}
\vspace{-2mm} 
\label{fig:census}
\end{subfigure}
\vspace{-3.5mm}
\caption{ Comparison of Cumulative Regrets over eight different datasets. The solid line is the average performance over 50 random seeds, with the shaded area representation $\pm$ one standard error. \label{fig:regret}}
\vspace{-2.0mm}
\end{figure*}

\begin{figure*}
 \centering
 \includegraphics[width=0.8\textwidth]{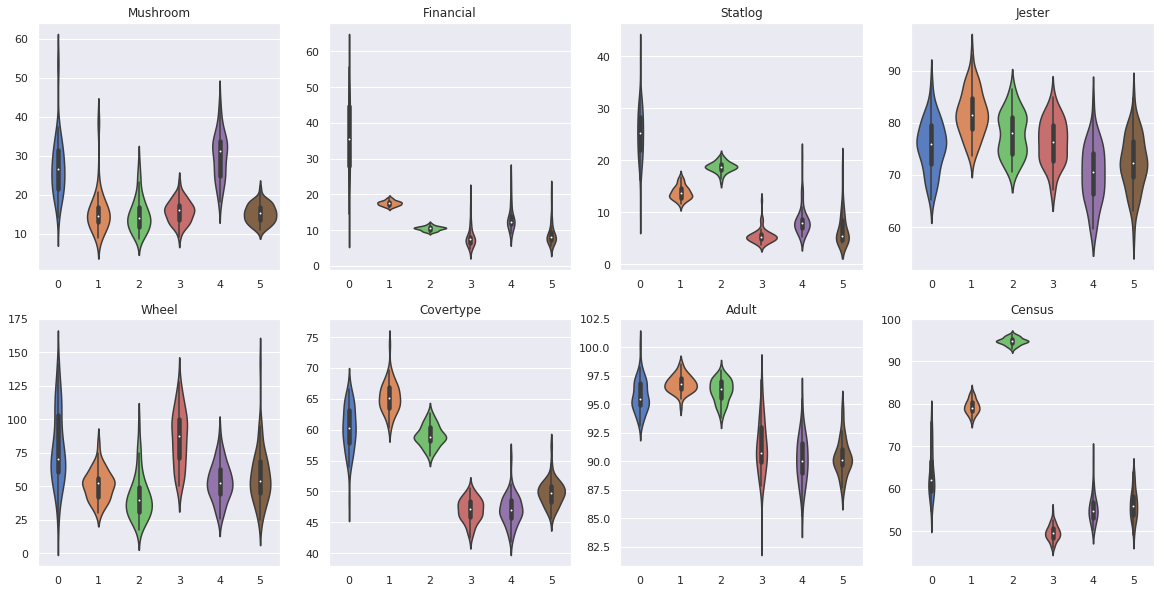}\vspace{-3.5mm}
 \caption{Boxplots of Normalized Cumulative Regrets on eight different datasets. The algorithms are ordered as: 0. BBB, 1. Neural Linear, 2. LinFullPost, 3. $\pi$-TS-DGF, 4. LU-Gauss, 5. LU-SIVI.}
 \label{fig:violin}
 \vspace{-1mm}
\end{figure*}

\subsection{Performance Comparison}

Following the same experimental settings of the contextual bandits benchmark in \citet{riquelme2018deep}, we evaluate the proposed LU-Gaussian and LU-SIVI and compare them to representative TS algorithms relying on global uncertainty. 
Neither LU-Gaussin nor LU-SIVI include noise injection to the global parameters, dropout layer, and bootstrapping techniques discussed in \citet{riquelme2018deep}. These techniques, designed to better capture global uncertainty to guide TS, can potentially be combined with LU-Gaussian and LU-SIVI to further improve their performance. We leave that for future study.

In Figure~\ref{fig:regret}, for each algorithm on a dataset, we 
plot the mean (a colored line) and standard %
error
(line shade) of its accumulative regret against the training step over 50 random trials. Using the performance of the Uniform algorithm, which uniformly at random chooses its actions from $\mathcal{A}$, as the reference, we show in Table \ref{tb:results} the normalized cumulative regrets and in Figure~\ref{fig:violin} their boxplots.

We first examine the performance of various global uncertainty guided TS algorithms, including BBB, LinFullPost, Nueral Linear, and $\pi$-TS-DGF. 
We find that BBB has the worst overall performance and exhibits large variance on its cumulative regrets across different random trials, suggesting that using a diagonal Gaussian variational distribution on the global parameters (neural network weights) leads to poor exploration. 
LinFullPost performs well in some datasets. E.g., it works very well on 
the Mushroom dataset, which is likely because using Gibbs sampling on global variables makes it simple to
move its reward distribution of eating a poisonous mushroom away from the bad local mode of~$+5$.
However, it clearly suffers from the lack of representation power on datasets that violate the linear assumption, such as Covertype, Census, Adult, and Statlog. 
Neural Linear involves feature extraction using a neural network and keeps the exact linear posterior updates for TS, but its representation learning relies heavily on the number of training steps that have been taken, which might be the reason for its relatively poor performance. Since $\pi$-TS-DGF of \citet{ruiyi2019ts} can provide better uncertainty estimation than BBB on the global parameters and more representation power than LinFullPost, it works quite well on some datasets and overall outperforms BBB and LinFullPost. 
Nevertheless, it does not perform that well on Mushroom and Jester and performs poorly on Wheel, which 
requires heavy exploration in order to minimize the cumulative regrets.

\begin{table*}[!t]
 \caption{
 Comparison of Normalized Cumulative Regret between various methods, with the normalization performed with respect to the Cumulative Regret of Uniform.
 {For each dataset, the same set of 50 random contextual sequences are used for all algorithms. For each algorithm on a given dataset, we report its mean and standard error over these 50 independent random trials.}
 }
 \label{tb:results}
 \begin{center}
 \resizebox{\textwidth}{!}{
 \begin{tabular}{lcccccccccc}
 \toprule
 \specialrule{0em}{1pt}{1pt}
 Algorithms & \small{Mean Rank} & \small{Mean Value} & Mushroom & Financial & Statlog & Jester & Wheel & Covertype & Adult & Census \\
 \specialrule{0em}{1pt}{1pt}
 \midrule
 Uniform & 7 & 100.00 & 100.00 \scriptsize{$\pm$ 4.95} & 100.00 \scriptsize{$\pm$ 11.63} & 100.00 \scriptsize{$\pm$ 1.04} & 100.00 \scriptsize{$\pm$ 7.67} & 100.00 \scriptsize{$\pm$ 5.73} & 100.00 \scriptsize{$\pm$ 0.88} & 100.00 \scriptsize{$\pm$ 0.50} & 100.00 \scriptsize{$\pm$ 0.63} \\
 \specialrule{0em}{1pt}{1pt}
 BBB & 4.75 & 56.53 & 26.08 \scriptsize{$\pm$ 7.34} & 34.31 \scriptsize{$\pm$ 9.72} & 25.93 \scriptsize{$\pm$ 4.62} & 75.58 \scriptsize{$\pm$ 4.06} & 71.12 \scriptsize{$\pm$ 18.51} & 60.37 \scriptsize{$\pm$ 3.91} & 95.64 \scriptsize{$\pm$ 1.29} & 63.15 \scriptsize{$\pm$ 4.29} \\
 \specialrule{0em}{1pt}{1pt}
 Neural Linear & 4.75 & 52.17 & 15.77 \scriptsize{$\pm$ 6.84} & 17.15 \scriptsize{$\pm$ 0.71} & 14.00 \scriptsize{$\pm$ 1.28} & 81.89 \scriptsize{$\pm$ 3.32} & 46.51 \scriptsize{$\pm$ 10.80} & 65.37 \scriptsize{$\pm$ 1.92} & 97.06 \scriptsize{$\pm$ 1.01} & 79.57 \scriptsize{$\pm$ 1.99} \\
 \specialrule{0em}{1pt}{1pt}
 LinFullPost & 3.75 & 51.15 & \textbf{13.66} \scriptsize{$\pm$ \textbf{3.77}} & 10.16 \scriptsize{$\pm$ 0.65} & 18.77 \scriptsize{$\pm$ 0.79} & 78.03 \scriptsize{$\pm$ 3.18} & \textbf{38.38} \scriptsize{$\pm$ \textbf{13.05}} & 58.83 \scriptsize{$\pm$ 1.62} & 96.05 \scriptsize{$\pm$ 0.97} & 95.26 \scriptsize{$\pm$ 0.77} \\
 \specialrule{0em}{1pt}{1pt}
 $\pi$-TS-DGF & \textbf{2.5} & 46.63 & 15.24 \scriptsize{$\pm$ 2.87} & \textbf{8.00} \scriptsize{$\pm$ \textbf{3.32}} & \textbf{6.10} \scriptsize{$\pm$ \textbf{2.75}} & 75.98 \scriptsize{$\pm$ 3.71} & 81.16 \scriptsize{$\pm$ 22.20} & \textbf{46.81} \scriptsize{$\pm$ \textbf{2.31}} & 90.77 \scriptsize{$\pm$ 2.00} & \textbf{49.02} \scriptsize{$\pm$ \textbf{1.96}} \\
 \specialrule{0em}{1pt}{1pt}
 LU-Gauss & 2.75 & 46.03 & 31.40 \scriptsize{$\pm$ 6.74} & 13.10 \scriptsize{$\pm$ 3.51} & 8.73 \scriptsize{$\pm$ 2.80} & \textbf{70.19} \scriptsize{$\pm$ \textbf{4.38}} & 52.90 \scriptsize{$\pm$ 15.54} & \textbf{47.33} \scriptsize{$\pm$ \textbf{2.56}} & \textbf{89.30} \scriptsize{$\pm$ \textbf{2.04}} & {55.28} \scriptsize{$\pm$ {2.83} } \\
 \specialrule{0em}{1pt}{1pt}
 LU-SIVI & \textbf{2.5} & \textbf{45.42} & 14.84 \scriptsize{$\pm$ 2.87} & 8.28 \scriptsize{$\pm$ 3.16} & 7.62 \scriptsize{$\pm$ 4.08} & 71.64 \scriptsize{$\pm$ 4.52} & 63.06 \scriptsize{$\pm$ 21.84} & 49.86 \scriptsize{$\pm$ 2.25} & \textbf{89.88} \scriptsize{$\pm$ \textbf{1.96}} & 58.18 \scriptsize{$\pm$ 5.45}\\
 \specialrule{0em}{1pt}{1pt}
 \bottomrule
 \end{tabular}
 }
 \end{center}\vspace{-1mm}
 \caption{{Analogous table to Table~\ref{tb:results} for Ablation Study. }
 }
 \label{tb:abl}
 \begin{center}
 \resizebox{\textwidth}{!}{
 \begin{tabular}{lcccccccccc}
 \toprule
 \specialrule{0em}{1pt}{1pt}
 Algorithms & \small{Mean Rank} & \small{Mean Value} & Mushroom & Financial & Statlog & Jester & Wheel & Covertype & Adult & Census \\
 \specialrule{0em}{1pt}{1pt}
 \midrule
 Uniform & 5 & 100.00 & 100.00 \scriptsize{$\pm$ 4.95} & 100.00 \scriptsize{$\pm$ 11.63} & 100.00 \scriptsize{$\pm$ 1.04} & 100.00 \scriptsize{$\pm$ 7.67} & 100.00 \scriptsize{$\pm$ 5.73} & 100.00 \scriptsize{$\pm$ 0.88} & 100.00 \scriptsize{$\pm$ 0.50} & 100.00 \scriptsize{$\pm$ 0.63} \\
 \specialrule{0em}{1pt}{1pt}
 LU-Gauss & \textbf{2.375} & 46.03 & 31.40 \scriptsize{$\pm$ 6.74} & 13.10 \scriptsize{$\pm$ 3.51} & 8.73 \scriptsize{$\pm$ 2.80} & 70.19 \scriptsize{$\pm$ 4.38} & \textbf{52.90} \scriptsize{$\pm$ \textbf{15.54}} & \textbf{47.33} \scriptsize{$\pm$ \textbf{2.56}} & \textbf{89.30} \scriptsize{$\pm$ \textbf{2.04}} & \textbf{55.28} \scriptsize{$\pm$ \textbf{2.83} } \\
 \specialrule{0em}{1pt}{1pt}
 LU-Gauss-Ablation & 2.5 & 46.55 & 21.28 \scriptsize{$\pm$ 10.57} & 12.35 \scriptsize{$\pm$ 5.09} & 7.45 \scriptsize{$\pm$ 2.83} & \textbf{69.98} \scriptsize{$\pm$ \textbf{4.63}} & 64.03 \scriptsize{$\pm$ 14.14} & 49.85 \scriptsize{$\pm$ 3.12} & 91.10 \scriptsize{$\pm$ 2.11} & {56.37} \scriptsize{$\pm$ {2.71} } \\
 \specialrule{0em}{1pt}{1pt}
 LU-SIVI & 2.5 & \textbf{45.42} & \textbf{14.84} \scriptsize{$\pm$ \textbf{2.87}} & \textbf{8.28} \scriptsize{$\pm$ \textbf{3.16}} & 7.62 \scriptsize{$\pm$ 4.08} & 71.64 \scriptsize{$\pm$ 4.52} & 63.06 \scriptsize{$\pm$ 21.84} & 49.86 \scriptsize{$\pm$ 2.25} & 89.88 \scriptsize{$\pm$ 1.96} & 58.18 \scriptsize{$\pm$ 5.45}\\
 LU-SIVI-Ablation & 2.625 & 47.81 & 20.32 \scriptsize{$\pm$ 11.35} & 9.01 \scriptsize{$\pm$ 2.71} & \textbf{6.34} \scriptsize{$\pm$ \textbf{3.53}} & 71.10 \scriptsize{$\pm$ 4.44} & 79.33 \scriptsize{$\pm$ 23.50} & 49.90 \scriptsize{$\pm$ 2.18} & 92.40 \scriptsize{$\pm$ 2.17} & 54.07 \scriptsize{$\pm$ 3.68}\\
 \specialrule{0em}{1pt}{1pt}
 \bottomrule
 \end{tabular}
 }
 \end{center}\vspace{-1mm}
 \caption{
 Details of the benchmark datasets and comparison of runtime (in seconds) between various methods. The reported runtime values are approximated with a single run.
 }
 \label{tb:runtime}
 \begin{center}
 \resizebox{0.7\textwidth}{!}{
 \begin{tabular}{clcccccccccc}
 \toprule
 \specialrule{0em}{1pt}{1pt}
 & & Mushroom & Financial & Statlog & Jester & Wheel & Covertype & Adult & Census \\
 \specialrule{0em}{1pt}{1pt}
 \midrule
 \multirow{2}{5em}{Dataset Information} & Context dimension & 22 & 21 & 16 & 32 & 2 & 54 & 94 & 389 \\
 & Number of actions & 2 & 8 & 7 & 8 & 5 & 7 & 14 & 9 \\
 \midrule
 \multirow{7}{4em}{Algorithms} & Uniform & 0.04 & 0.04 & 0.04 & 0.04 & 0.04 & 0.04 & 0.04 & 0.04 \\
 \specialrule{0em}{1pt}{1pt}
 & BBB & 13 & 13 & 12 & 13 & 12 & 13 & 13 & 15 \\
 \specialrule{0em}{1pt}{1pt}
 & Neural Linear & 22 & 35 & 33 & 35 & 28 & 33 & 50 & 40 \\
 \specialrule{0em}{1pt}{1pt}
 & LinFullPost & 14 & 7 & 4 & 9 & 3 & 13 & 48 & 710 \\
 \specialrule{0em}{1pt}{1pt}
 & $\pi$-TS-DGF & 270 & 107 & 82 & 128 & 62 & 168 & 246 & 750 \\
 \specialrule{0em}{1pt}{1pt}
 & LU-Gauss & 18 & 18 & 18 & 18 & 18 & 18 & 19 & 21 \\
 \specialrule{0em}{1pt}{1pt}
 & LU-SIVI & 32 & 30 & 30 & 30 & 30 & 31 & 31 & 40 \\
 \specialrule{0em}{1pt}{1pt}
 \bottomrule
 \end{tabular}
 }
 \end{center}
 \vspace{-2mm}
\end{table*}

We then examine the performance of the proposed LU guided TS algorithms, including LU-Gauss and LU-SIVI. As shown by Figures~\ref{fig:regret} and \ref{fig:violin} and Table \ref{tb:results}, LU-Gauss in general performs well, except that it provides poor performance on Mushroom, which, %
as analyzed in detail in Section \ref{sec:exploratory}, is likely because the diagonal Gaussian variational distribution is not flexible enough to encourage sufficient exploration. To be more specific, as in Figures~\ref{fig:singlemushroom}(b) and \ref{fig:manymushroom}(b), it 
prevents the sampled mean rewards of eating some poisonous mushrooms from moving away from a bad local density mode. By contrast, LU-SIVI performs well across all eight benchmark datasets. In particular, for Mushroom that LU-Guass performs poorly on, as shown in Figures~\ref{fig:singlemushroom}(d) and \ref{fig:manymushroom}(d), LU-SIVI places most of its sampled mean rewards of eating poisonous mushrooms clearly below zero. This can be explained by the improved ability of LU-SIVI in exploration due to its use of a semi-implicit variational distribution that is more flexible but remains simple to optimize.

To better compare the overall performance of different algorithms, as shown in Table~\ref{tb:results}, for each algorithm, we follow \citet{riquelme2018deep} and \citet{shengyang2019fv} to compute both the Mean Rank and Mean Value of normalized cumulative regrets over all eight benchmark datasets. The Mean Rank and Mean Value suggest that LU-SIVI has the best overall performance, followed by LU-Gauss and $\pi$-TS-DGF. In Appendix D, we further compare various algorithms in term of the Simple Regret metric \citep{riquelme2018deep}.

\subsection{ Ablation Study}
We introduce an ablation study to demonstrate the effectiveness of using local uncertainty. In the ablation baseline, everything stays the same as LU-Gauss/SIVI except that $\zv \sim q(\zv)$ becomes a global variable that no longer depends on $\xv_t$ in $q$, and $\text{KL} (q(\zv) || p(\zv))$ will be the single KL term shared by all observed data and hence needs to be appropriately rescaled in the objective function for stochastic variational inference. Specifically, in our framework, we change $\zv$ from via local uncertainty to via global uncertainty by replacing the input $\xv$ of $q(\zv \given \xv)$ with a constant vector of ones, which leads to $\zv$ shared by all data. 
We show the results of ablation baselines in Table \ref{tb:abl}. In general, LU-Gauss outperforms LU-Gauss-Ablation and LU-SIVI outperforms LU-SIVI-Ablation, suggesting the advantages of using local uncertainty over global uncertainty in our applications.

\subsection{Runtime comparison}
We report the time cost based on an Nvidia 1080-TI GPU.
Note when the contextual vector dimension is low, both LU-Gauss and LU-SIVI take more time to run than 
LinFullPost.
For example, on Mushroom whose contextual vector dimension is 22, LU-Gauss and LU-SIVI take about 18 and 30 seconds for 2000 training steps, while LinFullPost and Neural Linear take about 15 and 22 seconds. However, the computational complexity of LinFullPost increases cubically with the dimension of the contextual vector, due to the need to perform both Cholesky decomposition of the covariance matrix and matrix inversion when sampling the regression coefficient vector $\betav$. {Neural Linear involves a global latent layer $\zv$ with a fixed dimension to do runtime control.} In large contextual dimension case, for example, on Census whose contextual vector dimension is 389, it takes LU-Gauss and LU-SIVI about 20 and 40 seconds, respectively, to run 2000 training steps, while it takes LinFullPost and Neural Linear 710 and 40 seconds, respectively. For $\pi$-TS-DGF, it takes about 270 seconds on Mushroom and 750 seconds on Census, {due to its large number of particles.} Details of runtime comparison are shown in Table \ref{tb:runtime}.

\section{Conclusion}

To address the problem of contextual bandits, we propose Thompson sampling (TS) guided by local uncertainty (LU). This new TS framework models the reward distribution given the context using a latent variable model, and uses a pre-posterior contextual variational distribution to 
approximately capture the uncertainty of the local latent variable, whose true posterior depends on both the observed context and the reward that is yet to be observed. 
Under this framework, we introduce both LU-Gauss, which uses a diagonal Gaussian contextual variational distribution, 
and LU-SIVI, which uses a semi-implicit one. 
In comparison to LU-Gauss,
LU-SIVI has a more flexible variational distribution that enhances its ability of exploration, leading to improved performance on datasets with complex reward distributions. 
{Experimental results on eight different contextual bandit datasets demonstrate that both LU-Gauss and LU-SIVI well model the uncertainty and provide good performance, exhibiting reliability and robustness across all datasets. 
In particular, both LU-Gauss and LU-SIVI perform competitively to the particle-interactive TS algorithm of \citet{ruiyi2019ts}, the current state-of-the-art method, but have clearly lower computational complexity. The improved expressiveness, robustness, and computational complexity is the reason for the proposed local uncertainty guided TS method to claim a spot among the state-of-the-art.}
An interesting topic for future research is to investigate how to integrate both global parameter uncertainty and local latent variable uncertainty under TS to achieve further improved performance.

\section*{Acknowledgements}
The authors thank the anonymous reviewers, whose invaluable comments and suggestions have helped us to improve the paper.
This research was supported in part by Award IIS1812699 from the U.S. National Science Foundation. The authors acknowledge the Texas Advanced Computing Center (TACC) at The University of Texas at Austin for providing HPC resources that have contributed to the research results reported within this paper. URL: \url{http://www.tacc.utexas.edu}

\bibliography{References052016.bib,LUTS.bib}

\begin{thebibliography}{41}
\providecommand{\natexlab}[1]{#1}
\providecommand{\url}[1]{\texttt{#1}}
\expandafter\ifx\csname urlstyle\endcsname\relax
  \providecommand{\doi}[1]{doi: #1}\else
  \providecommand{\doi}{doi: \begingroup \urlstyle{rm}\Url}\fi

\bibitem[Agrawal \& Goyal(2012)Agrawal and Goyal]{agrawal2012analysis}
Agrawal, S. and Goyal, N.
\newblock Analysis of {T}hompson sampling for the multi-armed bandit problem.
\newblock In \emph{Conference on Learning Theory}, pp.\  39--1, 2012.

\bibitem[Agrawal \& Goyal(2013)Agrawal and Goyal]{agrawal2013ts}
Agrawal, S. and Goyal, N.
\newblock Thompson sampling for contextual bandits with linear payoffs.
\newblock In \emph{International Conference on Machine Learning}, 2013.

\bibitem[Arulkumaran et~al.(2017)Arulkumaran, Deisenroth, Brundage, and
  Bharath]{arulkumaran2017deep}
Arulkumaran, K., Deisenroth, M.~P., Brundage, M., and Bharath, A.~A.
\newblock Deep reinforcement learning: {A} brief survey.
\newblock \emph{IEEE Signal Processing Magazine}, 34\penalty0 (6):\penalty0
  26--38, 2017.

\bibitem[Auer(2002)]{auer2002uc}
Auer, P.
\newblock Using confidence bounds for exploitation-exploration trade-offs.
\newblock In \emph{Journal of Machine Learning Research}, 2002.

\bibitem[Bishop(2006)]{bishop2006pattern}
Bishop, C.~M.
\newblock \emph{Pattern Recognition and Machine Learning}.
\newblock Springer, 2006.

\bibitem[Bishop \& Tipping(2000)Bishop and Tipping]{bishop2000vr}
Bishop, C.~M. and Tipping, M.~E.
\newblock Variational relevance vector machines.
\newblock \emph{UAI}, pp.\  46--53, 2000.

\bibitem[Blei et~al.(2017)Blei, Kucukelbir, and McAuliffe]{blei2017variational}
Blei, D.~M., Kucukelbir, A., and McAuliffe, J.~D.
\newblock Variational inference: {A} review for statisticians.
\newblock \emph{Journal of the American Statistical Association}, 112\penalty0
  (518):\penalty0 859--877, 2017.

\bibitem[Blundell et~al.(2015)Blundell, Cornebise, Kavukcuoglu, and
  Wierstra]{charles2015bbb}
Blundell, C., Cornebise, J., Kavukcuoglu, K., and Wierstra, D.
\newblock Weight uncertainty in neural networks.
\newblock In \emph{International Conference on Learning Representations}, 2015.

\bibitem[Cesa-Bianchi et~al.(2017)Cesa-Bianchi, Gentile, Lugosi, and
  Neu]{cesa2017be}
Cesa-Bianchi, N., Gentile, C., Lugosi, G., and Neu, G.
\newblock Boltzmann exploration done right.
\newblock In \emph{Advances in Neural Information Processing Systems}, 2017.

\bibitem[Chapelle \& Li(2011)Chapelle and Li]{chapelle2011empirical}
Chapelle, O. and Li, L.
\newblock An empirical evaluation of {T}hompson sampling.
\newblock In \emph{Advances in neural information processing systems}, pp.\
  2249--2257, 2011.

\bibitem[Djallel \& Irina(2019)Djallel and Irina]{djallel2019}
Djallel, B. and Irina, R.
\newblock A survey on practical applications of multi-armed and contextual
  bandits.
\newblock \emph{arXiv preprint arXiv:1904.10040}, 2019.

\bibitem[Fortunato et~al.(2017)Fortunato, Gheshlaghi~Azar, Piot, Menick,
  Osband, Graves, Mnih, Munos, Hassabis, Pietquin, Blundell, and
  Legg]{Fortunato2017nn}
Fortunato, M., Gheshlaghi~Azar, M., Piot, B., Menick, J., Osband, I., Graves,
  A., Mnih, V., Munos, R., Hassabis, D., Pietquin, O., Blundell, C., and Legg,
  S.
\newblock Noisy networks for exploration.
\newblock In \emph{arXiv:1706.10295}, 2017.

\bibitem[Fran{\c{c}}ois-Lavet et~al.(2018)Fran{\c{c}}ois-Lavet, Henderson,
  Islam, Bellemare, Pineau, et~al.]{franccois2018introduction}
Fran{\c{c}}ois-Lavet, V., Henderson, P., Islam, R., Bellemare, M.~G., Pineau,
  J., et~al.
\newblock An introduction to deep reinforcement learning.
\newblock \emph{Foundations and Trends{\textregistered} in Machine Learning},
  11\penalty0 (3-4):\penalty0 219--354, 2018.

\bibitem[Gal \& Ghahramani(2016)Gal and Ghahramani]{gal2016da}
Gal, Y. and Ghahramani, Z.
\newblock Dropout as a bayesian approximation: Representing model uncertainty
  in deep learning.
\newblock In \emph{International Conference on Machine Learning}, 2016.

\bibitem[Graves(2011)]{graves2011practical}
Graves, A.
\newblock Practical variational inference for neural networks.
\newblock In \emph{Advances in Neural Information Processing Systems}, pp.\
  2348--2356, 2011.

\bibitem[Hern{\'a}ndez-Lobato \& Adams(2015)Hern{\'a}ndez-Lobato and
  Adams]{hernandez2015probabilistic}
Hern{\'a}ndez-Lobato, J.~M. and Adams, R.
\newblock Probabilistic backpropagation for scalable learning of {B}ayesian
  neural networks.
\newblock In \emph{International Conference on Machine Learning}, pp.\
  1861--1869, 2015.

\bibitem[Hinton \& Van~Camp(1993)Hinton and Van~Camp]{hinton1993keeping}
Hinton, G. and Van~Camp, D.
\newblock Keeping neural networks simple by minimizing the description length
  of the weights.
\newblock In \emph{in Proc. of the 6th Ann. ACM Conf. on Computational Learning
  Theory}. Citeseer, 1993.

\bibitem[Husz{\'a}r(2017)]{huszar2017variational}
Husz{\'a}r, F.
\newblock Variational inference using implicit distributions.
\newblock \emph{arXiv preprint arXiv:1702.08235}, 2017.

\bibitem[Jordan et~al.(1999)Jordan, Ghahramani, Jaakkola, and
  Saul]{jordan1999introduction}
Jordan, M.~I., Ghahramani, Z., Jaakkola, T.~S., and Saul, L.~K.
\newblock An introduction to variational methods for graphical models.
\newblock \emph{Machine learning}, 37\penalty0 (2):\penalty0 183--233, 1999.

\bibitem[Kingma \& Welling(2013)Kingma and Welling]{kingma2013vae}
Kingma, D.~P. and Welling, M.
\newblock Auto-encoding variational {B}ayes.
\newblock \emph{arXiv preprint arXiv:1312.6114}, 2013.

\bibitem[Li et~al.(2016)Li, Chen, Carlson, and Carin]{li2016psg}
Li, C., Chen, C., Carlson, D., and Carin, L.
\newblock Preconditioned stochastic gradient langevin dynamics for deep neural
  networks.
\newblock In \emph{Association for the Advancement of Artificial Intelligence},
  2016.

\bibitem[Maal{\o}e et~al.(2016)Maal{\o}e, S{\o}nderby, S{\o}nderby, and
  Winther]{maaloe2016auxiliary}
Maal{\o}e, L., S{\o}nderby, C.~K., S{\o}nderby, S.~K., and Winther, O.
\newblock Auxiliary deep generative models.
\newblock In \emph{ICML}, pp.\  1445--1453, 2016.

\bibitem[Mandt et~al.(2016)Mandt, Hoffman, and Blei]{mandt2016ava}
Mandt, S., Hoffman, M.~D., and Blei, D.~M.
\newblock A variational analysis of stochastic gradient algorithms.
\newblock In \emph{International Conference on Machine Learning}, 2016.

\bibitem[Mnih et~al.(2013)Mnih, Kavukcuoglu, Silver, Graves, Antonoglou,
  Wierstra, and Riedmiller]{mnih2013playing}
Mnih, V., Kavukcuoglu, K., Silver, D., Graves, A., Antonoglou, I., Wierstra,
  D., and Riedmiller, M.
\newblock Playing {A}tari with deep reinforcement learning.
\newblock \emph{arXiv preprint arXiv:1312.5602}, 2013.

\bibitem[Mnih et~al.(2015)Mnih, Kavukcuoglu, Silver, Rusu, Veness, Bellemare,
  Graves, Riedmiller, Fidjeland, Ostrovski, et~al.]{mnih2015human}
Mnih, V., Kavukcuoglu, K., Silver, D., Rusu, A.~A., Veness, J., Bellemare,
  M.~G., Graves, A., Riedmiller, M., Fidjeland, A.~K., Ostrovski, G., et~al.
\newblock Human-level control through deep reinforcement learning.
\newblock \emph{Nature}, 518\penalty0 (7540):\penalty0 529, 2015.

\bibitem[Mnih et~al.(2016)Mnih, Badia, Mirza, Graves, Lillicrap, Harley,
  Silver, and Kavukcuoglu]{mnih2016asynchronous}
Mnih, V., Badia, A.~P., Mirza, M., Graves, A., Lillicrap, T., Harley, T.,
  Silver, D., and Kavukcuoglu, K.
\newblock Asynchronous methods for deep reinforcement learning.
\newblock In \emph{International Conference on Machine Learning}, pp.\
  1928--1937, 2016.

\bibitem[Molchanov et~al.(2019)Molchanov, Kharitonov, Sobolev, and
  Vetrov]{molchanov2019doubly}
Molchanov, D., Kharitonov, V., Sobolev, A., and Vetrov, D.
\newblock Doubly semi-implicit variational inference.
\newblock In \emph{The 22nd International Conference on Artificial Intelligence
  and Statistics}, pp.\  2593--2602, 2019.

\bibitem[Neal(2012)]{neal2012bayesian}
Neal, R.~M.
\newblock \emph{Bayesian learning for neural networks}, volume 118.
\newblock Springer Science \& Business Media, 2012.

\bibitem[Osband et~al.(2016)Osband, Blundell, Pritzel, and
  Van~Roy]{osband2016de}
Osband, I., Blundell, C., Pritzel, A., and Van~Roy, B.
\newblock Deep exploration via bootstrapped {DQN}.
\newblock In \emph{Advances in Neural Information Processing Systems}, 2016.

\bibitem[Plappert et~al.(2017)Plappert, Houthooft, Dhariwal, Sidor, Y.~Chen,
  Chen, Asfour, Abbeel, and Andrychowicz]{plappert2017ps}
Plappert, M., Houthooft, R., Dhariwal, P., Sidor, S., Y.~Chen, R., Chen, X.,
  Asfour, T., Abbeel, P., and Andrychowicz, M.
\newblock Parameter space noise for exploration.
\newblock In \emph{arXiv:1706.01905}, 2017.

\bibitem[Ranganath et~al.(2016)Ranganath, Tran, and
  Blei]{ranganath2016hierarchical}
Ranganath, R., Tran, D., and Blei, D.
\newblock Hierarchical variational models.
\newblock In \emph{ICML}, pp.\  324--333, 2016.

\bibitem[Riquelme et~al.(2018)Riquelme, Tucker, and Snoek]{riquelme2018deep}
Riquelme, C., Tucker, G., and Snoek, J.
\newblock Deep {B}ayesian bandits showdown: {A}n empirical comparison of
  {B}ayesian deep networks for {T}hompson sampling.
\newblock In \emph{International Conference on Learning Representations}, 2018.
\newblock URL \url{https://openreview.net/forum?id=SyYe6k-CW}.

\bibitem[Russo et~al.(2018)Russo, Van~Roy, Kazerouni, Osband, and
  Wen]{russo2018tutorial}
Russo, D.~J., Van~Roy, B., Kazerouni, A., Osband, I., and Wen, Z.
\newblock A tutorial on {T}hompson sampling.
\newblock \emph{Foundations and Trends{\textregistered} in Machine Learning},
  11\penalty0 (1):\penalty0 1--96, 2018.

\bibitem[Sun et~al.(2019)Sun, Zhang, Shi, and Grosse]{shengyang2019fv}
Sun, S., Zhang, G., Shi, J., and Grosse, R.
\newblock Functional variational {B}ayesian neural networks.
\newblock In \emph{International Conference on Learning Representations}, 2019.

\bibitem[Sutton \& Barto(1998)Sutton and Barto]{sutton1998introduction}
Sutton, R.~S. and Barto, A.~G.
\newblock \emph{Introduction to reinforcement learning}, volume~2.
\newblock MIT press Cambridge, 1998.

\bibitem[Thompson(1933)]{thompson1933likelihood}
Thompson, W.~R.
\newblock On the likelihood that one unknown probability exceeds another in
  view of the evidence of two samples.
\newblock \emph{Biometrika}, 25\penalty0 (3/4):\penalty0 285--294, 1933.

\bibitem[Tran et~al.(2017)Tran, Ranganath, and Blei]{tran2017hierarchical}
Tran, D., Ranganath, R., and Blei, D.
\newblock Hierarchical implicit models and likelihood-free variational
  inference.
\newblock In \emph{Advances in Neural Information Processing Systems}, pp.\
  5523--5533, 2017.

\bibitem[Welling \& Teh(2011)Welling and Teh]{welling2011bayesian}
Welling, M. and Teh, Y.~W.
\newblock {B}ayesian learning via stochastic gradient {L}angevin dynamics.
\newblock In \emph{ICML}, pp.\  681--688, 2011.

\bibitem[Yin \& Zhou(2018)Yin and Zhou]{mingzhang2018sivi}
Yin, M. and Zhou, M.
\newblock Semi-implicit variational inference.
\newblock In \emph{International Conference on Machine Learning}, 2018.

\bibitem[Zhang et~al.(2020)Zhang, Chen, Tian, Wang, and
  Zhou]{Zhang2020Variational}
Zhang, H., Chen, B., Tian, L., Wang, Z., and Zhou, M.
\newblock Variational hetero-encoder randomized {GANs} for joint image-text
  modeling.
\newblock In \emph{International Conference on Learning Representations}, 2020.
\newblock URL \url{https://openreview.net/forum?id=H1x5wRVtvS}.

\bibitem[Zhang et~al.(2019)Zhang, Wen, Chen, and Carin]{ruiyi2019ts}
Zhang, R., Wen, Z., Chen, C., and Carin, L.
\newblock Scalable {T}hompson sampling via optimal transport.
\newblock In \emph{Artificial Intelligence and Statistics}, 2019.

\end{thebibliography}
\bibliographystyle{icml2020}

 \clearpage
 \appendix
 \onecolumn

 \begin{center}
 {\centering{\Large{\textbf{Thompson Sampling via Local Uncertainty: Appendix}}}}
 \end{center}

 \section{Model Details}
 In the implementation, we all use ReLU activation function, except for using the exponential link function for parameterizing the standard deviations of the Gaussian distributions. We set prior distribution $p(\zv_t)$ as a normal distribution with mean $\mathbf{0}$ and standard deviation $\sigma \Imat$, where $\sigma$ is a point estimation with initial value $\sigma=1.25$. Denote the latent dimension as $H$, the number of actions as $A$, and context dimension as $C$.

 For LU-Gauss: we set the latent dimension of $\zv$ as $H=50$. $\mathcal{T}_{\thetav}$ is a neural network composed of input layer $[\xv, \zv]$ in dimension $H + C$, one hidden layer [50], and output layer in dimension $A$. We use point estimate on $\Sigmamat_{\rv}$. 
 $\mathcal{T}_{\phiv_0}$ is a neural network composed of input layer $[\xv, \epsilonv]$ in dimension $C+C$, one hidden layer [100], and output layer in dimension 50, where $\epsilonv$ has the same dimension as context dimension. $\mathcal{T}_{\phiv_1}$ is a neural network composed of input layer in dimension 50, one hidden layer [50], and output layer in dimension $H$. $\mathcal{T}_{\phiv_2}$ is a neural network composed of input layer in dimension 50, one hidden layer [50], and output layer in dimension $H$.

 For LU-SIVI: we set the latent dimension of $\zv$ as $H=50$ and the number of noise $K=50$. $\mathcal{T}_{\thetav}$ is a neural network composed of input layer $[\xv, \zv]$ in dimension $H + C$, one hidden layer [50], and output layer in dimension $A$. We use point estimate on $\Sigmamat_{\rv}$. $\mathcal{T}_{\phiv_1}$ is a neural network composed of input layer $[\xv, \epsilonv]$ in dimension $C+C$, one hidden layer [100], and output layer in dimension $H$, where $\epsilonv$ has the same dimension as context dimension. $\mathcal{T}_{\phiv_2}$ is a neural network composed of input layer in dimension $C$, one hidden layer [50], and output layer in dimension $H$.

 \begin{figure}[H]
 \centering
 \includegraphics[scale=0.4]{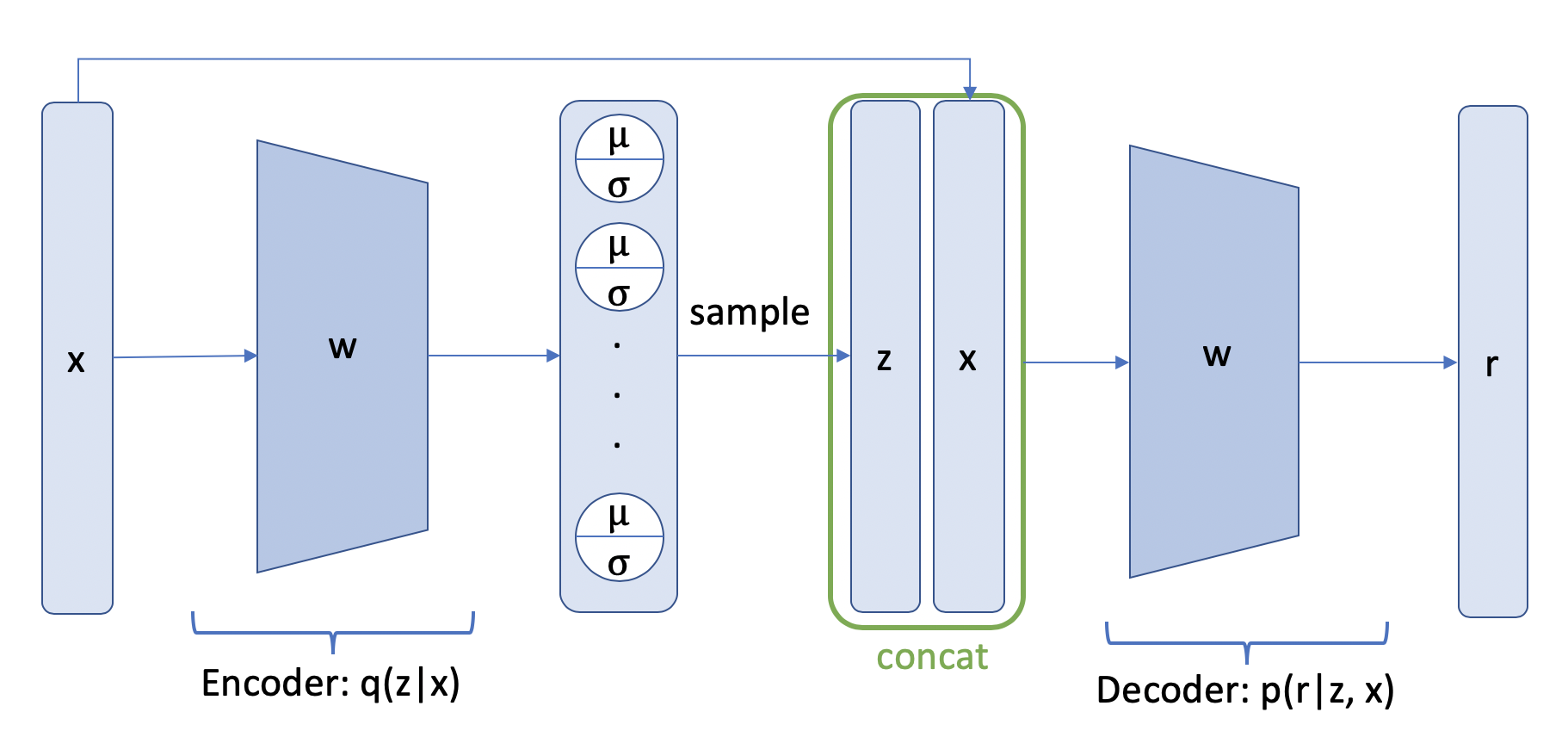}
 \caption{Model architecture}
 \label{fig:architecture}
 \end{figure}

 \section{Comparison with FBNNs}

 We have tried to make comparison with FBNNs of \citet{shengyang2019fv}.
 The code for FBNNs, however, 
 is too computationally expensive for us to run. 
 For example, the smallest FBNN model (1x50) used by \citet{shengyang2019fv} is already about 20 times slower than our proposed LU-Gauss and LU-SIVI, let along larger FBNN models. For this reason, we directly quote FBNNs' results reported in \citet{shengyang2019fv}, which were obtained based on as few as 10 random trials. Note the following results, which are quoted merely for reference, are not intended for a rigorous comparison as they are not obtained by averaging over the same set of 50 random sequences used by the other algorithms. 
 The Mean Value is 46.0 for FBNN ($1\times 50$), 47.0 for ($2\times 50$), 48.9 for ($3\times 50$), 45.3 for ($1\times 500$), 44.2 for ($2\times 500$), and 44.6 for ($3\times 500$).
 Their best Mean Value over the eight datasets is 44.2, from model FBNN (2 x 500), which is considerably slower to run in comparison to both LU-Gauss and LU-SIVI.

 \section{Details for TS models via Global Uncertainty}

 \textbf{Linear Method:} This method uses 
 Bayesian linear regression with closed-form Gibbs sampling update equations, which relies on the posterior distributions of regression coefficients for TS updates and maintains computational efficiency due to the use of conjugate priors. 
 This method assumes that at time $t$, the reward $y_t$ of an action given contextual input $\xv_t$ is generated as $y_t = \xv_t^T \betav + \epsilon_t$, 
 where
 $\betav$ is the vector of regression coefficients and $\epsilon_t \sim \cN (0, \sigma^2)$ is the noise. Note to avoid cluttered notation, we omit the action index. This method places a normal prior on $\betav$ and inverse gamma prior on $\sigma^2$. At time $t$, given $\xv_t$ and the current random sample of $\betav$, it takes the best action under TS and receives reward $y_t$; with $\xv_{1:t}$ and $y_{1:t}$, it samples $\sigma^2$ from its inverse gamma distributed conditional posterior, and then samples $\betav$ from its Gaussian distributed conditional posterior; it proceeds to the next time and repeats the same update scheme under TS.

 While this linear method accurately captures the posterior uncertainty of the global parameters $\betav$ and $\sigma^2$, its representation power is limited by both the linear mean and Gaussian distribution assumptions on reward $y$ given context $\xv$.
 In practice,
 the linear method often provides surprisingly competitive results, thanks to its ability to provide accurate uncertainty estimation.
 However, when its assumptions do not hold well in practice, such as when there are complex nonlinear dependencies between the rewards and contextual vectors, the linear method, even though with accurate posterior estimation, may not be able to converge to a good local optimal solution. Following \citet{riquelme2018deep}, we refer to this linear method as ``LinFullPost.'' 

 \textbf{Neural Linear:} To enhance the representation power of LinFullPost while maintaining closed-form posterior sampling, \citet{riquelme2018deep} propose the ``Neural Linear'' method, which feeds the representation of the last layer of a neural network as the covariates of a Bayesian linear regression model. It models the reward distribution of an action conditioning on $\xv$ as $y\sim\mathcal{N}(\betav^T\zv_{\xv},\sigma^2)$, 
 where $\zv_{\xv}$ is the output of the neural network given $\xv$ as the input. It separates representation learning and uncertainty estimation into two parts. The neural network part is responsible for finding a good representation of $\xv$, 
 while the Bayesian linear regression part is responsible for obtaining uncertainty estimation on the regression coefficient vector $\betav$, and making the decision on which action to choose under TS. The training for the two parts can be performed at different time-scales. It is reasonable to update the Bayesian linear regression part as soon as a new data arrives, while to update the neural network part only after collecting a number of new data points. 

 As Neural Linear transforms context $\xv$ into latent space $\zv$ via a deterministic neural network, the model uncertainty still all comes from sampling the global parameters $\betav$ and $\sigma$ from their posteriors under the Bayesian linear regression part. Hence, this method relies on the uncertainty of global model parameters to perform TS. 

 \textbf{Bayesian By Backprop (BBB):} This method uses variational inference to perform uncertainty estimation on the neural network weights \citep{charles2015bbb}. In order to exploit the reparameterization trick for tractable variational inference \citep{kingma2013vae}, it models the neural network weights with independent Gaussian distributions, whose means and variances become the network parameters to be optimized. However,
 the fully factorized mean-field variational inference used by BBB is well-known to have the tendency to underestimate posterior uncertainty \citep{jordan1999introduction,blei2017variational}.
 Moreover,
 it is also questionable whether the weight uncertainty can be effectively translated into reward uncertainty given context $\xv$ \citep{bishop2006pattern,shengyang2019fv}, especially considering that BBB makes both the independent and Gaussian assumptions on its network weights. For TS, underestimating uncertainty often leads to under exploration.
 As the neural network weights are shared across all observations, 
 BBB also relies on the uncertainty of global parameters to perform TS.

 \textbf{Particle-Interactive TS via Discrete Gradient Flow 
 ($\pi$-TS-DGF):} The $\pi$-TS-DGF 
 method of \citet{ruiyi2019ts} casts posterior approximation as a distribution optimization problem under the Wasserstein-gradient-flow framework. In this setting, posterior sampling in TS can be considered as a convex optimization problem on the space of probability measures.
 For tractability, it maintains a set of particles that interact with each other and evolve over time to approximate the posterior. 
 For the contextual bandit problem, each particle corresponds to a set of neural network weights, and the algorithm uniformly at random chooses one particle at each time and uses it as a posterior sample of the neural network weights. 
 A benefit of $\pi$-TS-DGF is that it imposes no explicit parametric assumption on the posterior distribution. 
 However, it faces an uneasy choice of setting the number of particles. Maintaining a large number of particles means training many sets of neural network weights at the same time, which is considerably expensive in computation, while a small number of particles might lead to bad uncertainty estimation due to inaccurate posterior approximation. The computational cost prevents $\pi$-TS-DGF from using large-size neural networks.
 Similar to BBB, $\pi$-TS-DGF also relies on the uncertainty of global parameters to perform exploration.

 \section{Results in Simple Regret Metric}
 We report the performance in terms of Simple Regret in Table \ref{tb:results_simple_regret}. Simple Regret is approximated by averaging the regrets over the last 500 steps.

 \begin{table*}[!h]
 \caption{Comparison of Normalized Simple Regret between various methods, with the normalization performed with respect to the Simple Regret of Uniform.
 {For each dataset, the same set of 50 random contextual sequences are used for all algorithms. For each algorithm on a given dataset, we report its mean and standard error over these 50 independent random trials.}
 }
 \label{tb:results_simple_regret}
 \begin{center}
 \resizebox{\textwidth}{!}{
 \begin{tabular}{lcccccccccc}
 \toprule
 \specialrule{0em}{1pt}{1pt}
 Algorithms & \small{Mean Rank} & \small{Mean Value} & Mushroom & Financial & Statlog & Jester & Wheel & Covertype & Adult & Census \\
 \specialrule{0em}{1pt}{1pt}
 \midrule
 Uniform & 7 & 100.00 & 100.00 \scriptsize{$\pm$ 9.70} & 100.00 \scriptsize{$\pm$ 12.62} & 100.00 \scriptsize{$\pm$ 1.76} & 100.00 \scriptsize{$\pm$ 8.66} & 100.00 \scriptsize{$\pm$ 13.20} & 100.00 \scriptsize{$\pm$ 1.88} & 100.00 \scriptsize{$\pm$ 1.38} & 100.00 \scriptsize{$\pm$ 1.48} \\
 \specialrule{0em}{1pt}{1pt}
 BBB & 5.125 & 52.77 & 19.36 \scriptsize{$\pm$ 8.58} & 30.25 \scriptsize{$\pm$ 19.04} & 30.91 \scriptsize{$\pm$ 8.80} & 71.54 \scriptsize{$\pm$ 5.28} & 60.62 \scriptsize{$\pm$ 25.12} & 62.32 \scriptsize{$\pm$ 7.74} & 95.73 \scriptsize{$\pm$ 2.63} & 51.41 \scriptsize{$\pm$ 6.31} \\
 \specialrule{0em}{1pt}{1pt}
 Neural Linear & 4.0 & 41.30 & 7.10 \scriptsize{$\pm$ 7.51} & 2.60 \scriptsize{$\pm$ 0.54} & 3.03 \scriptsize{$\pm$ 0.91} & 74.88 \scriptsize{$\pm$ 4.85} & 33.05 \scriptsize{$\pm$ 16.12} & 48.99 \scriptsize{$\pm$ 2.96} & 93.93 \scriptsize{$\pm$ 2.07} & 66.84 \scriptsize{$\pm$ 3.65} \\
 \specialrule{0em}{1pt}{1pt}
 LinFullPost & 3.125 & 41.06 & \textbf{4.14} \scriptsize{$\pm$ \textbf{3.06}} & \textbf{0.67} \scriptsize{$\pm$ \textbf{0.24}} & 11.28 \scriptsize{$\pm$ 1.90} & 69.64 \scriptsize{$\pm$ 4.87} & \textbf{20.62} \scriptsize{$\pm$ \textbf{13.68}} & 44.26 \scriptsize{$\pm$ 3.13} & 91.56 \scriptsize{$\pm$ 2.10} & 86.32 \scriptsize{$\pm$ 2.12} \\
 \specialrule{0em}{1pt}{1pt}
 $\pi$-TS-DGF & 3.125 & 42.15 & 6.37 \scriptsize{$\pm$ 2.91} & 2.73 \scriptsize{$\pm$ 3.31} & \textbf{1.01} \scriptsize{$\pm$ \textbf{1.79}} & 74.93 \scriptsize{$\pm$ 4.11} & 78.00 \scriptsize{$\pm$ 27.72} & 40.75 \scriptsize{$\pm$ 3.20} & 89.27 \scriptsize{$\pm$ 2.80} & \textbf{44.16} \scriptsize{$\pm$ \textbf{2.65}} \\
 \specialrule{0em}{1pt}{1pt}
 LU-Gauss & \textbf{2.625} & \textbf{38.83} & 27.79 \scriptsize{$\pm$ 11.04} & 3.73 \scriptsize{$\pm$ 2.96} & 1.99 \scriptsize{$\pm$ 3.13} & \textbf{67.91} \scriptsize{$\pm$ \textbf{5.38}} & 38.33 \scriptsize{$\pm$ 22.76} & \textbf{38.51} \scriptsize{$\pm$ \textbf{3.25}} & \textbf{85.85} \scriptsize{$\pm$ \textbf{3.33}} & {46.51} \scriptsize{$\pm$ {3.20} } \\
 \specialrule{0em}{1pt}{1pt}
 LU-SIVI & 3.0 & 40.08 & 7.07 \scriptsize{$\pm$ 4.06} & 1.96 \scriptsize{$\pm$ 3.09} & 3.70 \scriptsize{$\pm$ 7.07} & 68.43 \scriptsize{$\pm$ 5.27} & 55.64 \scriptsize{$\pm$ 25.79} & 42.76 \scriptsize{$\pm$ 3.60} & 87.08 \scriptsize{$\pm$ 2.74} & 54.02 \scriptsize{$\pm$ 11.00}\\
 \specialrule{0em}{1pt}{1pt}
 \bottomrule
 \end{tabular}
 }
 \end{center}\vspace{-4mm}
 \end{table*}

 \section{Algorithms}
 \small
 \begin{algorithm}[h]
 \caption{Vanilla Thompson Sampling}
 \label{alg:algov1}
 \begin{algorithmic}
 \STATE {\bfseries Input:} Prior distribution $p_0(\thetav)$.
 \STATE {\bfseries Output:} Fine-tuned posterior distribution $p_T(\thetav)$.
 \FOR{$t = 1, \dots, T$}
 {
 \STATE Observe context $\xv_t$.
 \STATE Sample parameters $\thetav_t \sim p_{t-1}(\thetav)$.
 \STATE Select action $a_t = \argmax_{a} f(\xv, a; \thetav_t)$.
 \STATE Observe reward $r_t$.
 \STATE Update posterior distribution $p_t(\thetav)$ with $(\xv_t, a_t, r_t)$.
 }
 \ENDFOR
 \end{algorithmic}
 \end{algorithm}
 \begin{algorithm*}[h]
 \caption{LU-Gauss: Thompson Sampling via Gaussian Local Uncertainty}
 \label{alg:algov2}
 \begin{algorithmic}
 \STATE {\bfseries Input:} {
 Likelihood $p(\rv_t \given \xv_t, \zv_t) = \cN(\muv_{\rv_t}, \Sigmamat_{\rv})$, prior $p(\zv)=\cN (\bf 0, \Sigmamat_{\zv})$, and variational distribution $q(\zv_t \given \xv_t) = \cN (\muv_{\zv_t}, \Sigmamat_{\zv_t})$; neural networks $\mathcal{T}_{\thetav}$, $\mathcal{T}_{\phiv_0}$, $\mathcal{T}_{\phiv_1}$, and $\mathcal{T}_{\phiv_2}$, $t_f=20$ (training frequency), $t_s=40$ (the number of mini-batches per training period).
 }
 \STATE {\bfseries Output:} {
 Inferred parameters $\Sigmamat_{\rv}$, $\thetav$, $\phiv_0$, $\phiv_1$, and $\phiv_2$.
 }
 \STATE Initialize $\Sigmamat_{\rv}$ 
 \STATE Initialize dataset $D_0 = \emptyset$ 
 \FOR{$t = 1, \dots, T$}
 {
 \STATE Observe context $\xv_t$
 \STATE $\muv_{\zv_t} = \mathcal{T}_{\phiv_1}(\hv)$, $\Sigmamat_{\zv_t} = \mathcal{T}_{\phiv_2}(\hv)$, $\hv = \mathcal{T}_{\phiv_0}(\xv_t)$
 \STATE Sample $\zv_t \sim \cN(\muv_{\zv_t}, \Sigmamat_{\zv_t})$
 \STATE $\muv_{\rv_t} = \mathcal{T}_{\thetav}([\xv_t, \zv_t])$ 
 \STATE Select action $a_t = \argmax_{a} \muv_{\rv_t}$ 
 \STATE Observe reward $r_t$ 
 \STATE $D_t = D_{t-1} \cup (\xv_t, a_t, r_t)$ 

 \IF{$t \text{ mod } t_f = 0$}
 {
 \FOR{$\text{iteration} = 1:t_s$}
 {
 \STATE Draw a minibatch data $\{(\xv_i, a_i, r_i)\}_{i=1}^N$
 \STATE Expand $r_i$ to vector $\rv_i$ with 0 at unobserved positions
 \STATE Create mask $\{ \mv_i\}_{i=1}^N$, where $\mv_i$ are zeros only except $\mv_i[a_i] = 1$
 \STATE Obtain the action space dimension as $|\cA|$ 
 \STATE $\hv_i = \mathcal{T}_{\phiv_0}(\xv_i)$, $\muv_{\zv_i} = \mathcal{T}_{\phiv_1}(\hv_i)$, $\Sigmamat_{\zv_i} = \mathcal{T}_{\phiv_2}(\hv_i)$ for $i=1:N$
 \STATE Let $\zv_i : = \muv_{\zv_i}+ \Sigmamat_{\zv_i}\odot\varepsilonv_i,~\varepsilonv_i\sim \mathcal N(\bf 0,\Imat)$ for $i=1:N$, where $\odot$ denotes element-wise product
 \STATE $\muv_{\rv_i} = \mathcal{T}_{\thetav}([\xv_i, \zv_i])$ for $i=1:N$
 \STATE Update $\Sigmamat_{\rv}$, $\thetav$, $\phiv_0$, $\phiv_1$, and $\phiv_2$ by using the gradients of $$\frac{1}{N} \sum_{i=1}^N \left[ \log p(\rv_i \given \xv_i, \zv_i) \cdot \mv_i \cdot |\cA| + \log \frac{p(\zv_i)}{q(\zv_i \given \xv_i)} \right] $$
 }
 \ENDFOR
 }
 \ENDIF
 }
 \ENDFOR

 \end{algorithmic}
 \end{algorithm*}

 \begin{algorithm*}[tb]
 \caption{LU-SIVI: Thompson Sampling with Semi-Implicit Local Uncertainty}
 \label{alg:algov3}
 \begin{algorithmic}
 \STATE {\bfseries Input:}
 {
 Likelihood $p(\rv_t \given \xv_t, \zv_t) = \cN (\muv_{\rv_t}, \Sigmamat_{\rv})$; prior $p(\zv)=\cN (\bf 0, \Sigmamat_{\zv})$; explicit variational distribution $q( \zv_t \given \xv_t) = \cN ( \psiv_t, \Sigmamat_{\zv_t})$ with reparameterization $\zv_t = 
 \muv_{\zv_t} + \Sigmamat_{\zv_t} \odot \varepsilonv_t$, ${\varepsilonv}_t \sim
 \mathcal{N}(\bf 0,\Imat)$; selected random noise distribution $q(\epsilonv)$; neural network $\mathcal{T}_{\thetav}$, $\mathcal{T}_{\phiv_1}$, and $\mathcal{T}_{\phiv_2}$; $t_f=20$ (training frequency), $t_s=40$ (the number of mini-batches per training period)
 }
 \STATE {\bfseries Output:}
 {
 Fine-tuned parameters $\Sigmamat_{\rv}, \thetav, \phiv_1, \phiv_2$.
 }

 \STATE Initialize parameters $\Sigmamat_{\rv}, \thetav, \phiv_1, \phiv_2$.
 \STATE Initialize dataset $D_0 = \emptyset$
 \FOR{$t = 1, \dots, T$}
 {
 \STATE Observe new context $\xv_t$

 \STATE $\psiv_t = \mathcal T_{\phiv_1}([\xv_t, \epsilonv_t]) $, where $\epsilonv_t \sim q(\epsilonv)$

 \STATE $\Sigmamat_{\zv_t} = \mathcal T_{\phiv_2}(\xv_t)$
 \STATE Sample $\zv_t \sim \cN ( \psiv_t, \Sigmamat_{\zv_t})$
 \STATE $\muv_{\rv_t} = \mathcal{T}_{\thetav}([\xv_t, \zv_t])$
 \STATE Select $a_t = \argmax \muv_{\rv_t}$
 \STATE Observe reward $r_t$
 \STATE $D_t = D_{t-1} \cup (\xv_t, a_t, r_t)$

 \IF{$t \text{ mod } t_f = 0$}
 {
 \FOR{$\text{iteration} = 1:t_s$}
 {
 \STATE Draw a minibatch data $\{(\xv_i, a_i, r_i)\}_{i=1}^N$
 \STATE Expand $r_i$ to vector $\rv_i$ with 0 at unobserved positions
 \STATE Create mask $\{ \mv_i\}_{i=1}^N$, where $\mv_i$ are zeros only except $\mv_i[a_i] = 1$
 \STATE Obtain the action space dimension as $|\cA|$
 \STATE Let $\psiv_i^{(k)}: =\mathcal T_{\phiv_1}([\xv_i, \epsilonv_i^{(k)}]),~\epsilonv_i^{(k)} \sim q(\epsilonv)$ for $k=0:K$
 \STATE Compute $ \Sigmamat_{\zv_i} =\mathcal T_{\phiv_2}(\xv_i)$
 \STATE Let $\zv_i: 
 = \psiv_i^{(0)}+\Sigmamat_{\zv_i}\odot \varepsilonv_i $,~~$\varepsilonv_i \sim \mathcal{N}(\bf 0, \Imat)$
 \STATE Compute $\muv_{\rv_i} = \mathcal{T}_{\thetav}([\xv_i, \zv_i])$

 \STATE Update $\Sigmamat_{\rv}, \thetav, \phiv_1, \phiv_2$ by using the gradients of: $$\frac{1}{N} \sum_{i=1}^N 
 \left[ \log p(\rv_i \given \muv_{\rv_i}, \Sigmamat_{\rv}) \cdot \mv_i \cdot |\cA| + \frac{\log p(\zv_i) } {\log \frac{1}{K+1} \sum_{k=0}^{K} q( \zv_i \given \psiv_i^{(k)}, \Sigmamat_{\zv_i})} \right] $$
 }
 \ENDFOR
 }
 \ENDIF
 }
 \ENDFOR
 \end{algorithmic}
 \end{algorithm*}
 \clearpage

\end{document}